\documentclass[11pt]{article}

\usepackage[preprint]{acl}

\usepackage{times}
\usepackage{latexsym}

\usepackage[T1]{fontenc}

\usepackage[utf8]{inputenc}

\usepackage{microtype}

\usepackage{inconsolata}

\usepackage{graphicx}

\usepackage{microtype}
\usepackage{booktabs,arydshln}
\usepackage{array}
\usepackage{ragged2e}
 \usepackage{amsmath}
 \usepackage{threeparttable}
 \usepackage{multicol}
 \usepackage{multirow}
 \usepackage{ccicons}
 \usepackage{chngcntr}

\makeatletter
\def\adl@drawiv#1#2#3{%
        \hskip.5\tabcolsep
        \xleaders#3{#2.5\@tempdimb #1{1}#2.5\@tempdimb}%
                #2\z@ plus1fil minus1fil\relax
        \hskip.5\tabcolsep}
\newcommand{\cdashlinelr}[1]{%
  \noalign{\vskip\aboverulesep
           \global\let\@dashdrawstore\adl@draw
           \global\let\adl@draw\adl@drawiv}
  \cdashline{#1}
  \noalign{\global\let\adl@draw\@dashdrawstore
           \vskip\belowrulesep}}
\makeatother

\makeatletter
\def\adl@drawiv#1#2#3{%
  \hskip.5\tabcolsep
  {\color[gray]{0.7} 
  \xleaders#3{#2.5\@tempdimb #1{1}#2.5\@tempdimb}%
      #2\z@ plus1fil minus1fil\relax}%
  \hskip.5\tabcolsep}
\newcommand{\cdashlinelrg}[1]{%
  \noalign{\vskip\aboverulesep
           \global\let\@dashdrawstore\adl@draw
           \global\let\adl@draw\adl@drawiv}
  \cdashline{#1}
  \noalign{\global\let\adl@draw\@dashdrawstore
           \vskip\belowrulesep}}
\makeatother

\usepackage{pifont}
\usepackage{fontawesome5}

\definecolor{richgreen}{RGB}{102, 204, 102}
\definecolor{richred}{RGB}{255, 99, 71}

\definecolor{richorange}{RGB}{255, 165, 0}

\definecolor{ctxorange}{RGB}{217,95,2} 
\definecolor{ctxteal}{RGB}{27,158,119} 
\definecolor{coolgray}{RGB}{160,160,170}

\definecolor{slateblue}{RGB}{100,140,200}

\definecolor{augpurple}{RGB}{117,107,177}

\usepackage[most]{tcolorbox}   
\usepackage{siunitx}           
\usepackage{colortbl} 

\newcolumntype{R}{p{0.05cm}} 

\newcolumntype{P}[1]{>{\RaggedRight\arraybackslash}p{#1}}
\usepackage{makecell}
\newcolumntype{C}[1]{>{\centering\arraybackslash}p{#1}}
\newcolumntype{D}[1]{>{\raggedright\arraybackslash}p{#1}}

\usepackage{amsmath}
\usepackage{amsfonts}
\usepackage{inconsolata}

\usepackage{graphicx}

\usepackage{blindtext}

\usepackage{afterpage}

\usepackage[capitalize]{cleveref}
\crefname{figure}{Figure}{Figures}
\crefname{table}{Table}{Tables}
\crefname{appendix}{Appendix}{Appendices}

\usepackage{todonotes}
\usepackage{soul}

\definecolor{TodoColor}{rgb}{1,0.7,0.6}

\usepackage{booktabs}   
\usepackage{colortbl}   
\usepackage[table]{xcolor} 
\usepackage[table,HTML]{xcolor}

\usepackage{enumitem}
\usepackage{placeins}

\usepackage{xstring}
\usepackage{seqsplit}

\newcommand{\capitalfirst}[1]{%
    \StrLeft{#1}{1}[\firstletter]%
    \StrGobbleLeft{#1}{1}[\restofword]%
    \MakeUppercase{\firstletter}\restofword%
}

\newcommand{\capitalizehyphenated}[1]{%
    \StrCut{#1}{-}{\firstpart}{\restpart}%
    \capitalfirst{\firstpart}%
    \IfStrEq{\restpart}{}{}{ \capitalizehyphenated{\restpart}}%
}

\newcommand{\hfmodel}[1]{%
    \StrBehind{#1}{/}[\hfmodelshortname]%
    \href{https://huggingface.co/#1}{\capitalizehyphenated{\hfmodelshortname}}%
}
\newcommand{\hfdata}[1]{%
    \StrBehind{#1}{/}[\hfmodelshortname]%
    \href{https://huggingface.co/datasets/#1}{\capitalizehyphenated{\hfmodelshortname}}%
}

\usepackage{amssymb}
\usepackage{pifont}

\usepackage{tcolorbox}
\tcbuselibrary{skins}
\newtcolorbox{promptbox}[1]{
    colback=gray!8,
    colframe=gray!45,
    title=#1,
    fonttitle=\small\bfseries,
    fontupper=\small\ttfamily,
    top=4pt, bottom=4pt
}

\usepackage{url}
\usepackage{xurl}
\usepackage{subcaption}

\title{Why We Need Speech to Evaluate Speech Translation}

\author{
 \textbf{Maike Züfle\textsuperscript{1}} \quad
 \textbf{Danni Liu\textsuperscript{1}} \quad
 \textbf{Vilém Zouhar\textsuperscript{2}} \quad
 \textbf{Jan Niehues\textsuperscript{1}}
\\
\\
 \textsuperscript{1}Karlsruhe Institute of Technology\quad 
 \textsuperscript{2}ETH Zurich
\\
 \small{
   \{\href{mailto:maike.zuefle@kit.edu}{maike.zuefle}, \href{mailto:danni.liu@kit.edu}{danni.liu}\}\textcolor{blue!50!black}{@kit.edu}, \href{mailto:vzouhar@ethz.ch}{vzouhar.ethz.ch}
 }
}

\begin{document}
\maketitle
\begin{abstract}
Speech translation models are increasingly capable of preserving speech-specific information (e.g., speaker gender, prosody, and emphasis), yet evaluation metrics remain blind to such phenomena.
We meta-evaluate both text- and speech-based quality estimation metrics on two contrastive datasets targeting gender agreement and prosody, and find that both fall short, even when given direct access to the speech signal.
We then train SpeechCOMET, a family of quality estimation models with speech encoders, and evaluate a state-of-the-art SpeechLLM as a judge. 
Both match or exceed text-based COMET on standard quality estimation, but neither consistently assesses speech-specific phenomena.
We identify three causes:
(1) speech-specific features are not reliably preserved in current encoders,
(2) models tend to ignore the speech source signal,
and (3) quality estimation training data contains too few relevant examples.
We release all models and code, and argue that progress requires dedicated speech-specific training data and models that genuinely condition on speech.
\end{abstract}

\section{Introduction}\label{sec:introduction}

\begin{figure}
    \centering
    \includegraphics[width=1.0\linewidth]{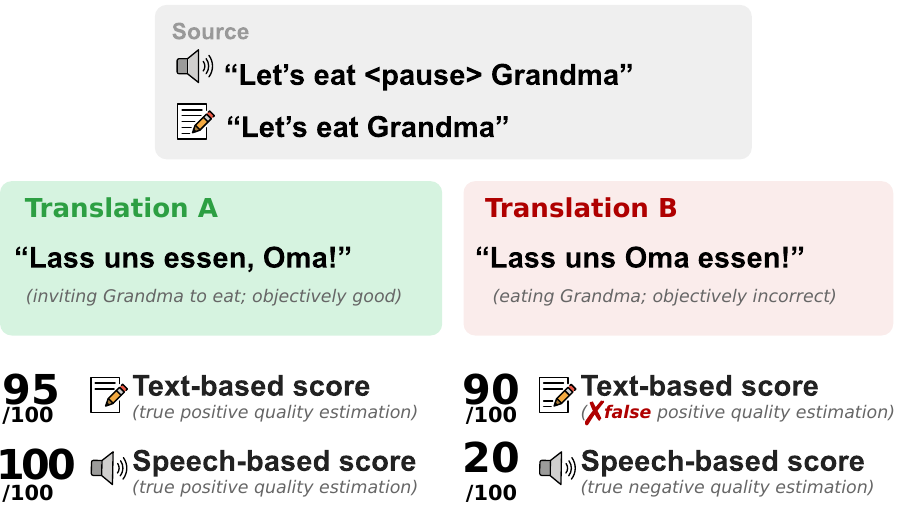}
    \caption{Speech is necessary to correctly evaluate speech translation quality: the prosodic break in the source speech disambiguates two translations.}
    \label{fig:intro_fig}
    \vspace{-3mm}
\end{figure}

Speech translation is evolving from cascaded pipelines, which first transcribe speech before translating the text, to end-to-end models~\citep{papi2026hearingtranslateeffectivenessspeech} that process speech directly and can, in principle, preserve speech-specific information such as speaker gender, prosody, and emphasis. The correct translation of an utterance can depend on such cues: the same words carry different meanings depending on the speech, as shown in \cref{fig:intro_fig}. 

Although cases where the speech signal is decisive may represent only a small fraction of all translations, the quality of speech translation has improved substantially recently \citep{agostinelli-etal-2025-findings} to the point where such phenomena can no longer be neglected. With the rise of Speech Large Language Models (SpeechLLMs; \citealp{xu2025qwen25omnitechnicalreport, microsoft2025phi4minitechnicalreportcompact}) and end-to-end ST models \citep{communication2023seamlessmultilingualexpressivestreaming, pmlr-v202-radford23a, ambilduke-etal-2025-tower}, recent work started to evaluate speech translation models on speaker gender agreement~\citep{zufle-niehues-2025-contrastive, conti2026voicebiascoreferenceinterpretability}, emotion \citep{papi2026hearingtranslateeffectivenessspeech} and prosodic phenomena \citep{Li_Zhao_Zhang_Su_Wei_Zhang_Chen_2026}, using dedicated contrastive benchmarks such as MuST-SHE~\citep{bentivogli-etal-2020-gender}, EmotionTalk \citep{sun2025emotiontalkinteractivechinesemultimodal} and ContraProST~\citep{tsiamas-etal-2024-speech}.

Yet dedicated quality estimation (QE) metrics for assessing these phenomena are missing. 
Current evaluations
rely on text-based QE models such as COMETKiwi~\citep{rei-etal-2022-comet}, which are blind to speech-specific phenomena by construction (\cref{fig:intro_fig}). While speech-aware metrics such as BLASER~\citep{dale-costa-jussa-2024-blaser} and SpeechQE~\citep{han-etal-2024-speechqe} have been proposed, they have only been evaluated on standard correlation with human scores, leaving open whether they can actually capture speech-specific phenomena.

We investigate whether existing and new metrics can close this gap. We show that existing speech-aware metrics, including BLASER and SpeechQE, are not sensitive to speech-specific phenomena despite receiving speech as input. We then introduce and train SpeechCOMET, a family of QE models with SONAR \citep{Duquenne:2023:sonar_arxiv} and Whisper \citep{pmlr-v202-radford23a} encoders, and additionally evaluate a state-of-the-art SpeechLLM as a judge. SpeechCOMET matches text-based COMETKiwi on general QE performance, and SpeechLLM exceeds it. However, neither SpeechCOMET nor SpeechLLM reliably assesses speech-specific phenomena.
We find three complementary explanations: (1) speech-specific features may not be preserved in the encoder representations at all; (2) models tend to ignore the speech source and focus on the translation; and (3)  even when features are present in the encoder, the QE training signal, which contains very few speech-specific examples, is insufficient to teach the model to use them.

Our contributions are threefold: (i) we benchmark state-of-the-art metrics on speech-specific evaluation; (ii) we train and evaluate new speech-aware QE models; and (iii) we provide hypotheses and evidence of why current models fail.

We release the trained QE models and code\footnote{\href{https://github.com/MaikeZuefle/speechCOMET}{https://github.com/MaikeZuefle/speechCOMET}}, and argue that progress in speech translation evaluation requires moving beyond text-based proxies and investing in speech-specific benchmarks and data.

\section{Current Evaluation of Speech Translation}\label{sec:existing}

We review existing approaches to speech translation quality  estimation (QE).

\subsection{Current Approaches}
\paragraph{Text-Based Metrics.}
The dominant approach to evaluating speech translation quality relies on text-based QE models. For shared tasks and in recent research, \hfmodel{Unbabel/WMT22-COMETKiwi-DA}~\citep{rei-etal-2022-comet} and its reference-based variant have been the standard metrics to date~\citep{agostinelli-etal-2025-findings, papi2026hearingtranslateeffectivenessspeech}, applied to human reference transcripts as source input. Two practical challenges arise in this setting. First, human transcripts are rarely available. Alternatives such as automatic speech recognition (ASR) transcripts and back-translations have been explored by \citet{sara-source-aware}, showing that ASR transcripts are more effective than back-translations when word error rate is low.
Second, beyond source representation, standard QE models assume perfectly segmented inputs, whereas speech translation usually relies on automatic segmentation~\citep{amrhein-haddow-2022-dont, agostinelli-etal-2025-findings}.
\hfmodel{zouharvi/COMET-Partial}~\citep{zouhar2025earlyexitinstantconfidencetranslation} addresses this by training on segmentation-noisy data. Yet none of these adaptations have speech access.
\begin{table*}[t]
\centering
\footnotesize
\setlength{\tabcolsep}{3pt}
\begin{tabular}{ll|rrrc|rrrc|rrr|rrr}
\toprule
 & & \multicolumn{8}{c|}{\textbf{IWSLT dev}} & \multicolumn{3}{c|}{\textbf{MuST-SHE}} & \multicolumn{3}{c}{\textbf{ContraProST}} \\
 & & \multicolumn{4}{c|}{Segment $\tau_b$ (\%)} & \multicolumn{4}{c|}{System SPA (\%)} & \multicolumn{3}{c|}{PA (\%)} & \multicolumn{3}{c}{PA (\%)} \\
\cmidrule(lr){3-6}\cmidrule(lr){7-10}\cmidrule(lr){11-13}\cmidrule(lr){14-16}
 & & \textbf{de} & \textbf{zh} & \textbf{avg} & \multicolumn{1}{c|}{\textbf{ASR src}} & \textbf{de} & \textbf{zh} & \textbf{avg} & \multicolumn{1}{c|}{\textbf{ASR src}} & \textbf{es} & \textbf{fr} & \textbf{it} & \textbf{de} & \textbf{es} & \textbf{ja} \\
\midrule
\multirow{2}{*}{\textsc{Text}} & \textsc{COMET-Partial} & \cellcolor[RGB]{255,255,255}11.2 & \cellcolor[RGB]{255,255,255}12.7 & \cellcolor[RGB]{255,255,255}11.9 & \cellcolor[RGB]{203,221,238}\scriptsize{-0.3} & \cellcolor[RGB]{255,255,255}43.1 & \cellcolor[RGB]{255,255,255}67.0 & \cellcolor[RGB]{255,255,255}55.0 & \cellcolor[RGB]{156,190,223}\scriptsize{+4.7} & \cellcolor[RGB]{253,240,229}52.4 & \cellcolor[RGB]{252,216,188}55.7 & \cellcolor[RGB]{253,230,212}51.8 & \cellcolor[RGB]{255,255,255}50.0 & \cellcolor[RGB]{255,255,255}50.0 & \cellcolor[RGB]{255,255,255}50.0 \\
 & \textsc{COMETKiwi} & \cellcolor[RGB]{183,224,223}\textbf{32.8} & \cellcolor[RGB]{183,224,223}\textbf{36.4} & \cellcolor[RGB]{183,224,223}\textbf{34.6} & \cellcolor[RGB]{255,255,255}\scriptsize{-6.0} & \cellcolor[RGB]{183,224,223}\textbf{86.3} & \cellcolor[RGB]{183,224,223}\textbf{89.1} & \cellcolor[RGB]{183,224,223}\textbf{87.7} & \cellcolor[RGB]{206,223,239}\scriptsize{-0.7} & \cellcolor[RGB]{250,183,132}\textbf{61.5} & \cellcolor[RGB]{250,183,132}\textbf{60.6} & \cellcolor[RGB]{250,183,132}\textbf{55.2} & \cellcolor[RGB]{255,255,255}50.0 & \cellcolor[RGB]{255,255,255}50.0 & \cellcolor[RGB]{255,255,255}50.0 \\
\multirow{2}{*}{\textsc{Speech}} & \textsc{SpeechQE} & \cellcolor[RGB]{203,232,232}26.9 & \cellcolor[RGB]{194,229,228}32.7 & \cellcolor[RGB]{198,230,230}29.8 & -- & \cellcolor[RGB]{195,229,228}79.2 & \cellcolor[RGB]{241,249,248}71.3 & \cellcolor[RGB]{210,236,235}75.2 & -- & \cellcolor[RGB]{255,255,255}37.0 & \cellcolor[RGB]{255,255,255}36.0 & \cellcolor[RGB]{255,255,255}31.0 & \cellcolor[RGB]{255,255,255}26.1 & \cellcolor[RGB]{255,255,255}20.0 & \cellcolor[RGB]{255,255,255}38.2 \\
 & \textsc{BLASER} & \cellcolor[RGB]{219,239,239}22.0 & \cellcolor[RGB]{212,236,236}26.8 & \cellcolor[RGB]{215,238,237}24.4 & -- & \cellcolor[RGB]{184,224,224}85.5 & \cellcolor[RGB]{253,254,254}67.4 & \cellcolor[RGB]{208,234,234}76.5 & -- & \cellcolor[RGB]{254,242,233}52.0 & \cellcolor[RGB]{254,244,236}51.6 & \cellcolor[RGB]{253,231,214}51.7 & \cellcolor[RGB]{232,168,205}\textbf{51.0} & \cellcolor[RGB]{232,168,205}\textbf{51.5} & \cellcolor[RGB]{232,168,205}\textbf{51.1} \\
\bottomrule
\end{tabular}
\vspace{-2mm}
\caption{Results for transcript- and speech-based baseline metrics. Transcript-based metrics cannot evaluate speech-specific phenomena by design; speech-based metrics can access the signal but still fail on gender and prosody.}\label{tab:baselines}
\vspace{-2.5mm}
\end{table*}

\paragraph{Speech-Based Metrics.}
A complementary line of work proposes metrics that operate directly on speech, avoiding ASR dependence altogether.
BLASER~\citep{chen-etal-2023-blaser} and its successor \hfmodel{facebook/blaser-2.0-qe}~\citep{dale-costa-jussa-2024-blaser} encode the source and translation directly into a shared multimodal embedding space using SONAR encoders~\citep{Duquenne:2023:sonar_arxiv}, originally targeted at speech-to-speech translation but extended to speech-to-text.
\hfmodel{h-j-han/SpeechQE-TowerInstruct-7B-en2de}~\citep{han-etal-2024-speechqe} takes a dedicated approach, combining Whisper \citep{pmlr-v202-radford23a} with an LLM to estimate speech-to-text translation quality. 
However, BLASER and SpeechQE are evaluated only on correlation with human scores or reference-based metrics, with no evaluation on benchmarks targeting speech-specific phenomena. 

\vspace{1.5em}
\noindent
We now benchmark existing metrics on both standard correlation with human judgements and on datasets targeting speech-specific phenomena. We describe the datasets, metrics, and results in turn.

\subsection{Data}\label{subsec:test_data}
\paragraph{IWSLT 2026 Metrics dev set.}
We evaluate on the development set of the IWSLT 2026 Metrics Shared Task \hfdata{maikezu/IWSLT2026-metrics-shared-train-dev}  \citep{adelani-etal-2026-iwslt}, which is, to the best of our knowledge, the only speech translation dataset with human-annotated scores\footnote{The human annotations from IWSLT 2023 are also public, we use them as part of the train set in later experiments.}, newly released after the existing metrics were developed. It comprises ACL talks from IWSLT 2025 submissions for en-de and en-zh. The dev set contains 5.56K segments annotated with human direct assessment scores across four speech translation systems.

The source texts in this dataset are human-annotated transcripts, which are rarely available in practice. We therefore  transcribe the source speech with \hfmodel{openai/whisper-large-v3}~\citep{pmlr-v202-radford23a} at a WER of 
24.4\%, and evaluate text-based models
with both human- or ASR source texts, 
to evaluate potential degradations~\citep{sara-source-aware}.

Beyond correlation with human scores, we evaluate whether QE metrics are sensitive to speech-specific phenomena. We use two contrastive datasets that test linguistically-motivated distinctions only accessible from the speech signal:

\paragraph{MuST-SHE.}
With MuST-SHE~\citep{bentivogli-etal-2020-gender}, we evaluate whether metrics are sensitive to speaker gender agreement in speech translation. The dataset contains English speech translated to  French, Spanish, and Italian, which inflect grammatical forms according to the speaker's gender. Each of the 1612 evaluation pairs consists of a gender-correct translation and a gender-incorrect alternative. A metric that uses speech should prefer translations that match the speaker's gender.

\paragraph{ContraProST.}
ContraProST~\citep{tsiamas-etal-2024-speech} tests sensitivity to prosodic and paralinguistic features: correct translations are paired with prosodically incorrect alternatives across en$\to$\{de, es, ja\} (15,972 pairs), covering five phenomena: sentence stress, intonation, prosodic breaks, emotional prosody, and politeness. These distinctions are not recoverable from transcripts alone, making this a direct test of whether speech-based metrics capture information beyond text.

\subsection{Meta-Evaluation Metrics}
We adopt the meta-evaluation metrics from the IWSLT 2026 Metrics Shared Task\footnote{\href{https://iwslt.org/2026/metrics}{iwslt.org/2026/metrics}}, assessing our QE systems at both the segment and system level. For text-based models, we additionally evaluate the effect of using automatic transcripts as source inputs, which may contain ASR errors.

\paragraph{Segment-level correlation.}
For each source segment, we compute Kendall's $\tau_b$~\citep{kendall} between metric predictions and human direct assessment scores across the available system translations, 
and then average this across source segments. A higher $\tau_b$ indicates better agreement between the metric and human ranking of translations.

\paragraph{System-level correlation.}
At the system level, we evaluate the metric's ability to rank speech translation systems over a full dataset.
We use Soft Pairwise Accuracy \citep[SPA,][]{thompson-etal-2024-improving}, which measures agreement between human and automatic pairwise system rankings. 
SPA is 1 when the metric assigns identical pairwise confidence to every system comparison as human judgements.

\paragraph{Contrastive pairwise accuracy.}
For contrastive evaluation datasets such as MuST-SHE and ContraProST, no human scores are available. Instead, each source is paired with one correct and one incorrect translation. We measure whether the metric assigns a higher score to the correct translation:\vspace{-0.2cm}
\begin{align}
\text{PA} = \frac{\sum_{k=1}^{N} \mathbf{1}[\hat{y}_k^+ > \hat{y}_k^-]}{N}
\end{align}

where $N$ is the number of contrastive pairs, $\hat{y}_k^+$ is the metric score for the correct translation, and $\hat{y}_k^-$ is the score for the incorrect one. PA equals 1 when the metric always prefers the correct translation, and ties are counted as incorrect.

\subsection{Evaluating Existing Metrics}

We benchmark existing approaches on the datasets described above; results are shown in \cref{tab:baselines}. Transcript-based metrics outperform speech-based ones when given human reference transcripts, but degrade noticeably under ASR input: COMET drops from 32.8\% to 27.8\% at the segment level for en-de and from 36.4\% to 29.4\% for en-zh. With ASR transcripts, transcript-based and speech-based metrics perform comparably.

For evaluations targeting speech-specific phenomena, neither family performs well.
Transcript-based metrics score at chance on MuST-SHE and ContraProST by design, as they have no access to the speech signal. More surprisingly, speech-based metrics also do not perform well: 
despite receiving speech as input, SpeechQE and BLASER score at or below chance on both datasets.

\section{New Approaches to Speech\newline Translation Evaluation}\label{sec:experiments}

Having established that existing metrics are not sensitive to speech-specific phenomena, we investigate whether this limitation is architectural. We train SpeechCOMET, an adaptation of COMETKiwi with a speech encoder, and evaluate a state-of-the-art SpeechLLM as a judge to test whether either approach can capture speech-specific phenomena. 

\subsection{SpeechCOMET}
\paragraph{Architecture.}
Reference-free text-based QE models \citep{rei-etal-2022-comet} encode the source sentence and the hypothesis separately using a shared multilingual encoder, and feed
their representations into an MLP estimator. Concretely, let $\mathbf{s}_t \in \mathbb{R}^d$ and $\mathbf{h}_t \in \mathbb{R}^d$ denote the pooled source and hypothesis embeddings, respectively. The estimator receives the concatenation of four interaction features:
\begin{equation}
    \mathbf{e} = \bigl[\mathbf{h}_t;\, \mathbf{s}_t;\,
                       |\mathbf{h}_t - \mathbf{s}_t|;\,
                       \mathbf{h}_t \odot \mathbf{s}_t \bigr] \in \mathbb{R}^{4d},\label{eq:4way}
\end{equation}
and produces a scalar quality score $\hat{y} = \text{MLP}(\mathbf{e})$, where the MLP is a two-layer feed-forward network with Tanh activations.

In SpeechCOMET, we replace the text source encoder with a speech representation $\mathbf{s}_a \in \mathbb{R}^d$ that is projected to match the text encoder dimension $d$ via a learned linear layer. The hypothesis is still encoded by a text encoder, and the same four-way interaction is applied.
Two alternative fusion strategies are described and 
evaluated in \cref{app:harris_fusion}. They perform worse than the four-way interaction.

We further extend SpeechCOMET to take \emph{both} the source speech and source text as input. The two source representations are fused into a single embedding
$\mathbf{s} = f(\mathbf{s}_t, \mathbf{s}_a)$, where $f$ is one of three strategies: element-wise average, element-wise sum, or a learned linear projection applied to their concatenation $[\mathbf{s}_t;\, \mathbf{s}_a] \in \mathbb{R}^{2d}$. The fused embedding $\mathbf{s}$ then replaces $\mathbf{s}_t$ in \cref{eq:4way}.

\paragraph{Speech and Text Embedding Models.}
We encode the hypothesis text using \hfmodel{FacebookAI/XLM-RoBERTa-base}~\citep{roberta-base-2019}, a multilingual transformer model, choosing the base model over the larger InfoXLM-large~\citep{chi-etal-2021-infoxlm} used by COMETKiwi~\citep{rei-etal-2022-comet} for efficiency. To enable a direct comparison at equal encoder capacity, we additionally train our best model with InfoXLM-large (SpeechCOMET$^{\text{XL}}$). We also train a text-only baseline using the smaller encoder to isolate the contribution of speech input.

For source speech, we investigate two encoders. SONAR~\citep{Duquenne:2023:sonar_arxiv} is a multilingual speech encoder that produces fixed-size sentence-level embeddings.
\hfmodel{openai/whisper-large-v3}~\citep{pmlr-v202-radford23a} outputs a variable-length sequence of frame-level representations. For the latter, we use only the encoder component and evaluate two aggregation strategies: average pooling and attention-based pooling over the encoder output.

\paragraph{Training Details.}
For both speech encoders, we experiment with frozen and fine-tuned variants. For SONAR, we partially fine-tune by freezing the bottom four transformer layers and updating the remainder, motivated by probing studies showing that phonetic information is concentrated in the middle to upper-middle layers~\citep{which_sonar_layers_1, which_sonar_layers_2}. For Whisper, full fine-tuning is memory-prohibitive, so we instead use LoRA~\citep{lora-hu-2021} while keeping the base weights frozen. We train on a single NVIDIA A100-SXM4-40GB GPU. Hyperparameters are listed in Table~\ref{tab:speechcomet-hyperparams}.

\paragraph{Data.}
All models are trained on source-hypothesis pairs annotated with human direct assessment scores, where the source is either text or speech depending on the model variant.
The \hfdata{maikezu/IWSLT2026-metrics-shared-train-dev} dataset \citep{adelani-etal-2026-iwslt} provides paired source speech and source text alongside human quality scores, comprising 229 hours of audio across 33{,}721 samples. 
As the IWSLT training set is relatively small, we investigate whether augmenting with text-quality-score pairs from \hfdata{zouharvi/WMT-human-all} dataset~\citep{kocmi-etal-2025-findings} improves performance. The two sets differ in domain, as WMT covers text translation with gold segmentation, while IWSLT covers speech translation with automatic segmentation. WMT provides only the source text, so we synthesise speech for the 490{,}943 WMT samples using \hfmodel{hexgrad/Kokoro-82M} TTS to obtain 1{,}114 hours of paired audio-quality-score examples. Language pairs and further dataset details are provided in \cref{app:data}.
We continue to evaluate all models on the test sets described in \cref{subsec:test_data}.

\begin{table*}[t]
\centering
\footnotesize
\setlength{\tabcolsep}{3pt}
\begin{tabular}{ll|rrrc|rrrc|rrr|rrr}
\toprule
 & & \multicolumn{8}{c|}{\textbf{IWSLT dev}} & \multicolumn{3}{c|}{\textbf{MuST-SHE}} & \multicolumn{3}{c}{\textbf{ContraProST}} \\
 & & \multicolumn{4}{c|}{Segment $\tau_b$ (\%)} & \multicolumn{4}{c|}{System SPA (\%)} & \multicolumn{3}{c|}{PA (\%)} & \multicolumn{3}{c}{PA (\%)} \\
\cmidrule(lr){3-6}\cmidrule(lr){7-10}\cmidrule(lr){11-13}\cmidrule(lr){14-16}
 & & \textbf{de} & \textbf{zh} & \textbf{avg} & \multicolumn{1}{c|}{\textbf{ASR src}} & \textbf{de} & \textbf{zh} & \textbf{avg} & \multicolumn{1}{c|}{\textbf{ASR src}} & \textbf{es} & \textbf{fr} & \textbf{it} & \textbf{de} & \textbf{es} & \textbf{ja} \\
\midrule
\textsc{Text} & \textsc{COMETKiwi} & \cellcolor[RGB]{205,233,233}\textbf{32.8} & \cellcolor[RGB]{224,241,241}\textbf{36.4} & \cellcolor[RGB]{217,238,238}\textbf{34.6} & \cellcolor[RGB]{246,249,252}\scriptsize{-6.0} & \cellcolor[RGB]{196,229,229}\textbf{86.3} & \cellcolor[RGB]{183,224,223}\textbf{89.1} & \cellcolor[RGB]{185,225,224}\textbf{87.7} & \cellcolor[RGB]{208,224,240}\scriptsize{-0.7} & \cellcolor[RGB]{250,183,132}\textbf{61.5} & \cellcolor[RGB]{250,183,132}\textbf{60.6} & \cellcolor[RGB]{250,183,132}\textbf{55.2} & \cellcolor[RGB]{255,255,255}\textbf{50.0} & \cellcolor[RGB]{255,255,255}\textbf{50.0} & \cellcolor[RGB]{255,255,255}\textbf{50.0} \\
\midrule
\multicolumn{16}{l}{\textit{SpeechCOMET}} \\
\midrule
\multirow{2}{*}{\textsc{Text}} & COMETKiwi$_{\text{RoBERTa}}^{\text{WMT}}$ & \cellcolor[RGB]{244,250,250}20.3 & \cellcolor[RGB]{242,249,249}25.9 & \cellcolor[RGB]{243,249,249}23.1 & \cellcolor[RGB]{214,228,242}\scriptsize{-1.5} & \cellcolor[RGB]{255,255,255}44.7 & \cellcolor[RGB]{211,236,235}67.4 & \cellcolor[RGB]{255,255,255}56.0 & \cellcolor[RGB]{192,214,235}\scriptsize{+1.5} & \cellcolor[RGB]{254,247,242}51.2 & \cellcolor[RGB]{253,236,222}52.8 & \cellcolor[RGB]{254,253,252}50.1 & \cellcolor[RGB]{255,255,255}\textbf{50.0} & \cellcolor[RGB]{255,255,255}\textbf{50.0} & \cellcolor[RGB]{255,255,255}50.0 \\
 & COMETKiwi$_{\text{RoBERTa}}^{\text{IWSLT}}$ & \cellcolor[RGB]{244,250,250}20.2 & \cellcolor[RGB]{243,250,250}25.3 & \cellcolor[RGB]{243,250,250}22.8 & \cellcolor[RGB]{206,223,239}\scriptsize{-0.5} & \cellcolor[RGB]{219,239,239}69.9 & \cellcolor[RGB]{211,236,235}67.4 & \cellcolor[RGB]{227,243,242}68.7 & \cellcolor[RGB]{216,229,242}\scriptsize{-1.8} & \cellcolor[RGB]{253,237,225}52.8 & \cellcolor[RGB]{255,255,255}49.1 & \cellcolor[RGB]{253,227,207}52.0 & \cellcolor[RGB]{255,255,255}\textbf{50.0} & \cellcolor[RGB]{255,255,255}\textbf{50.0} & \cellcolor[RGB]{255,255,255}50.0 \\
\multirow{2}{*}{\textsc{Speech}} & SpeechCOMET\textsubscript{SONAR} & \cellcolor[RGB]{253,254,254}17.3 & \cellcolor[RGB]{247,251,251}22.9 & \cellcolor[RGB]{249,252,252}20.1 & -- & \cellcolor[RGB]{246,251,251}51.0 & \cellcolor[RGB]{206,234,233}\textbf{70.9} & \cellcolor[RGB]{244,250,250}61.0 & -- & \cellcolor[RGB]{253,235,221}53.1 & \cellcolor[RGB]{254,248,244}50.9 & \cellcolor[RGB]{254,242,233}50.9 & \cellcolor[RGB]{255,255,255}\textbf{50.0} & \cellcolor[RGB]{255,255,255}\textbf{50.0} & \cellcolor[RGB]{255,255,255}50.0 \\
 & SpeechCOMET\textsubscript{Whisper} & \cellcolor[RGB]{255,255,255}16.8 & \cellcolor[RGB]{255,255,255}18.7 & \cellcolor[RGB]{255,255,255}17.8 & -- & \cellcolor[RGB]{254,254,254}44.9 & \cellcolor[RGB]{209,235,235}68.5 & \cellcolor[RGB]{253,254,254}56.7 & -- & \cellcolor[RGB]{254,246,240}51.4 & \cellcolor[RGB]{254,250,246}50.7 & \cellcolor[RGB]{252,216,188}\textbf{52.8} & \cellcolor[RGB]{255,255,255}\textbf{50.0} & \cellcolor[RGB]{255,255,255}\textbf{50.0} & \cellcolor[RGB]{232,168,205}\textbf{50.1} \\
\textsc{Sp.+Txt} & SpeechCOMET & \cellcolor[RGB]{231,244,244}24.5 & \cellcolor[RGB]{240,248,248}27.1 & \cellcolor[RGB]{237,247,247}25.8 & \cellcolor[RGB]{230,238,247}\scriptsize{-3.8} & \cellcolor[RGB]{205,233,233}79.7 & \cellcolor[RGB]{211,236,235}67.4 & \cellcolor[RGB]{216,238,238}73.5 & \cellcolor[RGB]{213,227,241}\scriptsize{-1.4} & \cellcolor[RGB]{252,220,195}\textbf{55.6} & \cellcolor[RGB]{253,237,224}52.6 & \cellcolor[RGB]{252,222,198}52.4 & \cellcolor[RGB]{255,255,255}\textbf{50.0} & \cellcolor[RGB]{255,255,255}49.9 & \cellcolor[RGB]{255,255,255}50.0 \\
\midrule
\textsc{Sp.+Txt} & SpeechCOMET$^{\text{XL}}$ & \cellcolor[RGB]{205,233,233}\textbf{32.7} & \cellcolor[RGB]{225,242,241}\textbf{36.0} & \cellcolor[RGB]{218,239,238}\textbf{34.4} & \cellcolor[RGB]{246,249,252}\scriptsize{-6.0} & \cellcolor[RGB]{198,230,230}\textbf{85.2} & \cellcolor[RGB]{210,235,235}67.8 & \cellcolor[RGB]{210,235,235}\textbf{76.5} & \cellcolor[RGB]{156,190,223}\scriptsize{+6.5} & \cellcolor[RGB]{253,226,205}54.6 & \cellcolor[RGB]{251,201,162}\textbf{58.0} & \cellcolor[RGB]{253,239,228}51.1 & \cellcolor[RGB]{255,255,255}\textbf{50.0} & \cellcolor[RGB]{255,255,255}\textbf{50.0} & \cellcolor[RGB]{255,255,255}50.0 \\
\midrule
\multicolumn{16}{l}{\textit{SpeechLLM}} \\
\midrule
\multirow{2}{*}{\textsc{Text}} & SpeechLLM & \cellcolor[RGB]{203,233,232}33.2 & \cellcolor[RGB]{205,233,233}47.5 & \cellcolor[RGB]{204,233,233}40.4 & \cellcolor[RGB]{243,247,251}\scriptsize{-5.5} & \cellcolor[RGB]{183,224,223}\textbf{95.6} & \cellcolor[RGB]{255,255,255}32.6 & \cellcolor[RGB]{237,247,247}64.1 & \cellcolor[RGB]{188,211,233}\scriptsize{+2.1} & \cellcolor[RGB]{255,255,255}23.7 & \cellcolor[RGB]{255,255,255}17.3 & \cellcolor[RGB]{255,255,255}16.2 & \cellcolor[RGB]{255,255,255}23.8 & \cellcolor[RGB]{255,255,255}18.1 & \cellcolor[RGB]{255,255,255}31.1 \\
 & \quad+FT & \cellcolor[RGB]{184,224,224}39.3 & \cellcolor[RGB]{192,228,227}55.0 & \cellcolor[RGB]{189,227,226}47.1 & \cellcolor[RGB]{230,239,247}\scriptsize{-3.8} & \cellcolor[RGB]{186,225,225}93.4 & \cellcolor[RGB]{190,227,226}\textbf{83.7} & \cellcolor[RGB]{183,224,223}\textbf{88.5} & \cellcolor[RGB]{227,237,246}\scriptsize{-3.4} & \cellcolor[RGB]{255,255,255}14.0 & \cellcolor[RGB]{255,255,255}13.8 & \cellcolor[RGB]{255,255,255}15.4 & \cellcolor[RGB]{255,255,255}14.0 & \cellcolor[RGB]{255,255,255}14.0 & \cellcolor[RGB]{255,255,255}26.1 \\
\multirow{2}{*}{\textsc{Speech}} & SpeechLLM & \cellcolor[RGB]{224,242,241}26.5 & \cellcolor[RGB]{222,241,240}37.5 & \cellcolor[RGB]{223,241,241}32.0 & -- & \cellcolor[RGB]{194,229,228}87.5 & \cellcolor[RGB]{254,254,254}32.7 & \cellcolor[RGB]{246,251,251}60.1 & -- & \cellcolor[RGB]{255,255,255}38.0 & \cellcolor[RGB]{255,255,255}32.3 & \cellcolor[RGB]{255,255,255}32.4 & \cellcolor[RGB]{255,255,255}\textbf{32.2} & \cellcolor[RGB]{255,255,255}\textbf{28.4} & \cellcolor[RGB]{255,255,255}\textbf{33.1} \\
 & \quad+FT & \cellcolor[RGB]{202,232,231}33.8 & \cellcolor[RGB]{207,234,234}46.5 & \cellcolor[RGB]{205,233,233}40.1 & -- & \cellcolor[RGB]{191,227,227}90.2 & \cellcolor[RGB]{232,245,245}50.1 & \cellcolor[RGB]{224,241,241}70.1 & -- & \cellcolor[RGB]{255,255,255}35.9 & \cellcolor[RGB]{255,255,255}15.0 & \cellcolor[RGB]{255,255,255}20.7 & \cellcolor[RGB]{255,255,255}21.1 & \cellcolor[RGB]{255,255,255}21.9 & \cellcolor[RGB]{255,255,255}15.7 \\
\multirow{2}{*}{\textsc{Sp.+Txt}} & SpeechLLM & \cellcolor[RGB]{207,234,234}32.2 & \cellcolor[RGB]{211,236,235}44.1 & \cellcolor[RGB]{209,235,235}38.1 & \cellcolor[RGB]{236,242,248}\scriptsize{-4.5} & \cellcolor[RGB]{190,227,226}90.7 & \cellcolor[RGB]{255,255,255}32.6 & \cellcolor[RGB]{242,249,249}61.6 & \cellcolor[RGB]{207,223,239}\scriptsize{-0.6} & \cellcolor[RGB]{255,255,255}\textbf{43.1} & \cellcolor[RGB]{255,255,255}\textbf{37.1} & \cellcolor[RGB]{255,255,255}\textbf{37.8} & \cellcolor[RGB]{255,255,255}31.0 & \cellcolor[RGB]{255,255,255}26.1 & \cellcolor[RGB]{255,255,255}32.6 \\
 & \quad+FT & \cellcolor[RGB]{183,224,223}\textbf{39.7} & \cellcolor[RGB]{183,224,223}\textbf{60.1} & \cellcolor[RGB]{183,224,223}\textbf{49.9} & \cellcolor[RGB]{236,243,249}\scriptsize{-4.6} & \cellcolor[RGB]{192,228,227}89.1 & \cellcolor[RGB]{205,233,233}72.1 & \cellcolor[RGB]{201,231,231}80.6 & \cellcolor[RGB]{255,255,255}\scriptsize{-7.2} & \cellcolor[RGB]{255,255,255}8.2 & \cellcolor[RGB]{255,255,255}6.3 & \cellcolor[RGB]{255,255,255}6.8 & \cellcolor[RGB]{255,255,255}17.8 & \cellcolor[RGB]{255,255,255}8.8 & \cellcolor[RGB]{255,255,255}20.4 \\
\bottomrule
\end{tabular}
\vspace{-3mm}
\caption{Results for SpeechCOMET and SpeechLLM models. ASR src shows the change in average $\tau_b$/SPA when using ASR transcripts instead of human transcripts; speech-only models have no text input (--).}\label{tab:main_results_v2}
\vspace{-2.5mm}
\end{table*}

\paragraph{Model Variants.}
We train the following model variants. \textit{Text models} are trained on two datasets: WMT direct assessment data, which provides a large amount of training signal, and the IWSLT dataset using source texts. The latter enables a fair comparison with speech models, which are trained on IWSLT only. \textit{Speech models} use source speech encoded by SONAR or Whisper and are trained in frozen and fine-tuned encoder variants. \textit{Speech+text models} take both modalities as input and are trained with three fusion strategies (average, sum, projection). We additionally train speech+text models initialised from a text checkpoint, to investigate whether a strong text initialisation benefits multimodal models. Finally, we train \textit{TTS-augmented models} on WMT data with synthesised speech, as well as combined IWSLT+WMT-TTS variants, to study the effect of training data size and domain on speech model quality.

\subsection{SpeechLLM as Judge}
As a second approach we evaluate \hfmodel{Qwen/Qwen2.5-Omni-7B}~\citep{xu2025qwen25omnitechnicalreport}, a multimodal large language model capable of processing text, audio, and both modalities simultaneously, as a zero-shot and fine-tuned quality estimator. Unlike SpeechCOMET, which is trained specifically for QE, the LLM is prompted to produce a scalar quality score directly, without task-specific architectural adaptations.

We evaluate three input modalities, text only, speech only, and speech+text, mirroring the SpeechCOMET setup. In the \textit{zero-shot} setting, the model is prompted with each modality using a standard QE prompt as well as two task-specific prompt variants for MuST-SHE and ContraProST (see Figure~\ref{fig:speechllm_prompts}). In the \textit{fine-tuned} setting, we train one model per modality using LoRA adapters on the IWSLT dataset described earlier. Fine-tuning and inference are performed on a single NVIDIA A100-SXM4-40GB GPU. Hyperparameters and prompts are listed in \cref{app:speechllm}.

\section{Results}\label{sec:results}
Results for all model families are given in \cref{tab:main_results_v2}. A significance analysis and per-category results for ContraProST are given in \cref{subsec:significance}.

\subsection{SpeechCOMET Results}

\paragraph{Text-only: in-domain training data outweighs scale.}
We train two text-only SpeechCOMET variants: one on large-scale WMT text translation data (COMETKiwi$_{\text{RoBERTa}}^{\text{WMT}}$) and one on IWSLT speech translation system outputs (COMETKiwi$_{\text{RoBERTa}}^{\text{IWSLT}}$). Despite WMT being substantially larger, COMETKiwi$_{\text{RoBERTa}}^{\text{IWSLT}}$ achieves much higher system-level correlation (68.7 vs.\ 56.0 SPA avg), while segment-level performance is comparable (22.8 vs.\ 23.1 $\tau_b$ avg). As expected for text-only models, neither variant shows sensitivity to gender (MuST-SHE) or prosody (ContraProST).

\begin{figure*}[t]
  \centering
\includegraphics[width=\textwidth]{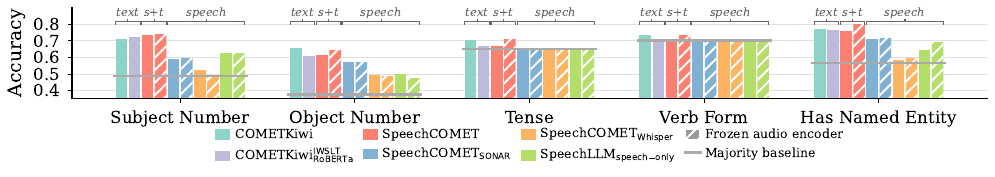}
\vspace{-8mm}
  \caption{Probing accuracy for linguistic features on the IWSLT 2026 Metrics dataset.   Scores averaged over 3 runs (std. $\le$~0.02).
  \textit{Text-aware} systems consistently outperform \textit{speech-only} systems.
  Fine-tuned variants (hatched) closely track their frozen counterparts (solid).
  SONAR-based SpeechCOMET outperforms Whisper-based variants.
}
\label{fig:linguistics_feat_probing}
\vspace{-4mm}
\end{figure*}

\paragraph{Speech-only: limited sensitivity to speech-specific phenomena.}
Next, we replace the text encoder with a speech encoder, using either SONAR or Whisper, to encode the source speech directly. For each encoder, we experiment with freezing vs.\ unfreezing the encoder during fine-tuning and with augmenting training data with a TTS-synthesized variant of WMT data (see \cref{sec:experiments}). The best-performing variant of each is reported in \cref{tab:main_results_v2}, with the full ablation in \cref{tab:tts_ablation} in \cref{app:training_strat}. Both encoders benefit from unfreezing, but only Whisper benefits from TTS pretraining. Unlike SONAR, Whisper is an ASR model with no native sentence-level representation and requires additional speech exposure to adapt.

The two speech models slightly underperform COMETKiwi$_{\text{RoBERTa}}^{\text{IWSLT}}$, which is trained on the same IWSLT data but with text as source, indicating that text remains a stronger signal than speech.  
In general, SpeechCOMET\textsubscript{SONAR} outperforms SpeechCOMET\textsubscript{Whisper} (20.1 vs.\ 17.8 $\tau_b$ avg) on the IWSLT test set. Neither model shows sensitivity to gender or prosody.

\paragraph{Speech+text: multimodal fusion improves correlation.}
We then combine both modalities by fusing the individual source embeddings. We ablate different fusion strategies and two initialisation schemes: training from scratch and initialising from an existing text SpeechCOMET checkpoint. We report the best model, sum fusion initialised from COMETKiwi$_{\text{RoBERTa}}^{\text{WMT}}$, in \cref{tab:main_results_v2}; the full ablation is in \cref{tab:orkney_ablation} in \cref{app:fusion_speech_text}. 

SpeechCOMET achieves 25.8 $\tau_b$ and 73.5 SPA avg, outperforming both unimodal variants. MuST-SHE and ContraProST performance remains at chance. Scaling the text encoder to InfoXLM-large (SpeechCOMET$^{\text{XL}}$), the same encoder as the off-the-shelf COMETKiwi uses, closes the gap to the COMETKiwi text baseline (34.4 vs.\ 34.6 $\tau_b$), suggesting that the performance gap of the smaller models relative to COMETKiwi is due to encoder capacity rather than training procedure. However, this gain is driven entirely by the stronger text encoder, as speech sensitivity remains unchanged.


\subsection{SpeechLLM as a Judge Results}
We now discuss the performance of the SpeechLLM. Generally, SpeechLLM outperforms all SpeechCOMET variants and the COMETKiwi text baseline on IWSLT, with the best model reaching 49.9 $\tau_b$ avg compared to 34.4 for SpeechCOMET$^{\text{XL}}$ and 34.6 for COMETKiwi (\cref{tab:main_results_v2}).

\paragraph{Text-only: performance improves with FT.}
The SpeechLLM with only text input achieves 40.4 $\tau_b$ avg without fine-tuning, and 47.1 with IWSLT fine-tuning, with similarly large gains at system level (64.1 to 88.5 SPA avg), outperforming the SpeechCOMET models. The below-chance PA scores on MuST-SHE and ContraProST are driven by ties: the model assigns identical scores to most pairs, and PA is exactly 50\% once ties are excluded.

\paragraph{Speech-only: speech input underperforms text.}
With speech input, SpeechLLM performs below its text-only counterpart (32.0 vs.\ 40.4 $\tau_b$ avg, 60.1 vs.\ 64.1 SPA avg), with the gap persisting after fine-tuning (40.1 vs.\ 47.1 $\tau_b$). Despite having access to the speech signal, MuST-SHE and ContraProST show the same tie-driven pattern as the text-only model, with no gender or prosody detection.

\paragraph{Speech+text: combining modalities yields best correlation but no speech sensitivity.}
Combining both modalities yields the strongest results: 49.9 $\tau_b$ avg with FT and 80.6  SPA avg with FT, surpassing all other models. Still, the model does not show sensitivity to speech-specific phenomena.

\paragraph{Specific prompts do not enhance speech sensitivity.}
Since speech sensitivity does not emerge from architecture or fine-tuning, we additionally test whether it can be elicited through task-specific prompting. However, instructing the model to attend to paralinguistic cues  yields no consistent improvement on MuST-SHE or ContraProST. We report these results in \cref{tab:speechllm_prompt} in \cref{app:results}.

\section{Analyses}\label{subsec:anaylsis}
Having shown that no current metric is sensitive to speech-specific phenomena, we test three complementary explanations: the features are 1) missing from encoder representations, 2) not induced during training, or 3) disregarded during QE scoring.

\subsection{Are features present in the encoders?}
We probe for source linguistic and acoustic features in \cref{fig:linguistics_feat_probing} and \ref{fig:mustshe_probing}.
For each system, we extract the source representations and train an MLP probe \cite{probing} to predict the target feature. For COMET and SpeechCOMET, we use either the encoder's fixed-size sentence embedding (SONAR, RoBERTa for text) or the utterance representation from the pooling layer learned during QE training (Whisper). For SpeechLLMs, we use the hidden state immediately following the prompt.
Details about the probing experiments are in \cref{app:probing_setup}.

\subsubsection{Probing Linguistic Features}
We probe five linguistic features: subject number, object number (singular, plural, none in both cases), tense (present, past, none), verb form (finite, infinite, participle, none), and the presence of named entities (binary).
\cref{fig:linguistics_feat_probing} shows three patterns across these probed linguistic features. 

\paragraph{Modality gap separates text-aware from speech-only systems.}
There is a clear gap between input modalities: systems that have access to text at the source, whether text-only (COMETKiwi variants) or speech-and-text combined (SpeechCOMET), achieve substantially higher probing accuracy than their speech-only counterparts.
This pattern holds consistently across all five features and suggests that a considerable amount of linguistically structured information present in the text signal is not recoverable from the audio signal alone, 
at least as encoded by the models studied here.

\paragraph{QE fine-tuning does not close the speech-text modality gap.}
Pairwise comparison of fine-tuned and frozen variants of the same architecture reveals nearly identical probing accuracies. 
The absence of improvement from fine-tuning indicates that the quality estimation training signal alone is insufficient to drive the encoder towards more linguistically informative representations of source speech, likely because standard QE training provides no examples that would enforce such features.

\paragraph{SONAR representations dominate Whisper in preserving linguistic features.}
SONAR provides a fixed-size utterance-level embedding pre-trained to capture semantic content, 
whereas Whisper's utterance representation is obtained via a pooling mechanism learned from scratch during quality estimation training. 
The fact that the off-the-shelf SONAR embedding outperforms the jointly trained Whisper pooling reinforces the conclusion that the quality estimation objective, in its current form, does not provide sufficient  signal for creating high-quality audio source representations.

\subsubsection{Probing Acoustic Features}
\begin{figure}[t!]
  \centering
\includegraphics[width=\linewidth,trim=0cm 0cm 0cm 0.1cm, clip]{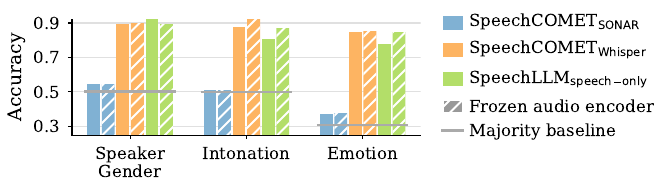}
  \caption{Probing accuracy for speaker gender on MuST-SHE and for intonation (question vs.\ statement) and emotion (angry, happy, neutral, surprised) on ContraProST.
  Scores averaged over three runs (std. $\le$~0.02). 
  SpeechCOMET\textsubscript{Whisper} and SpeechLLMs enable accurate recovery of acoustic features while SpeechCOMET\textsubscript{SONAR} does not.
  }
  \label{fig:mustshe_probing}
  \vspace{-2mm}
\end{figure}

Having probed for linguistic features, 
we now test whether acoustic and speaker-level features can be recovered from the source representations (\cref{fig:mustshe_probing}). 
We probe three features: speaker gender, intonation, and emotion. 
We focus on speech-only models, 
and defer the full set of text-only and speech-text models (\cref{tab:main_results_v2}) to \cref{app:additional_probing_results}.
These features are not recoverable from text alone, so the probe must rely on acoustic information.

Overall, the trend is consistent across the three probed features: 
Whisper representations and the SpeechLLM hidden states support accurate recovery, while SONAR representations do not.
The results have the following implications.

\paragraph{SONAR's semantic alignment may erase acoustic information.}
SpeechCOMET with the SONAR backbone, despite outperforming the Whisper variant on the earlier linguistic probes (\cref{fig:linguistics_feat_probing}), 
now performs at or just above the majority baseline on acoustic features. 
A plausible reason is that SONAR's training objective, 
which aligns sentences with the same meaning across languages, 
pulls together utterances that share meaning but differ acoustically. 
Features like speaker gender and intonation,
which are largely uncorrelated with a sentence's overall meaning,
would then be suppressed as a side effect.
This is consistent with \citet{fucci-etal-2025-different}, who find that in speech+MT architectures, gender information is well encoded in the speech encoder output but largely lost after the adapters that project speech into the MT model's text embedding space, suggesting that alignment to a text-semantic space, rather than any property of SONAR specifically, is what suppresses acoustic attributes.

\paragraph{Whisper and SpeechLLMs encode speaker gender but do not use it.}
For both Whisper and the SpeechLLMs, the probe recovers speaker gender with high accuracy, confirming that this information is present in the source representation. Yet neither is sensitive to acoustic inputs in its QE outputs (\cref{tab:main_results_v2}). 
Therefore, the feature is available but unused. 
Unlike SONAR, where the acoustic information appears to be erased from the audio representation, 
here the bottleneck lies not in the encoder but in the training signal, which does not reward attending to acoustic information.

\subsection{Source Reliance of the QE Models}
Automated metrics are notorious for focusing primarily on the translation only and disregarding the sources \citep{sun-etal-2020-estimating,zouhar-etal-2023-poor}.
To quantify how much the models rely on their source input when assigning QE scores, we replace every source with a randomly mismatched one and measure the drop in segment $\tau_b$ in \cref{tab:random_src}. 

\textsc{COMETKiwi} and the speech baseline \textsc{SpeechQE} show strong source sensitivity, dropping by $-24.9$ and $-22.5$ pp respectively, yet the latter is still not able to use speech-specific features (discussion in \cref{sec:discussion}).

\begin{table}[t]
\centering
\footnotesize
\setlength{\tabcolsep}{4pt}
\begin{tabular}{ll|rc}
\toprule
 \multicolumn{2}{r}{\textbf{Segment $\tau_b$ (\%)}} & \textbf{real src} & \textbf{$\Delta$ random} \\
\midrule
\multicolumn{4}{l}{\textit{Baselines}} \\
\midrule
\multirow{2}{*}{\textsc{Text}} & \textsc{COMET-PART.} & \cellcolor[RGB]{255,255,255}11.9 & \cellcolor[RGB]{249,251,253}-4.3 \\
 & \textsc{COMETKiwi} & \cellcolor[RGB]{212,236,236}34.6 & \cellcolor[RGB]{194,215,235}-24.9 \\
\multirow{2}{*}{\textsc{Speech}} & \textsc{SpeechQE} & \cellcolor[RGB]{221,240,240}29.8 & \cellcolor[RGB]{201,219,237}-22.5 \\
 & \textsc{BLASER} & \cellcolor[RGB]{231,244,244}24.4 & \cellcolor[RGB]{220,232,243}-15.4 \\
\midrule
\multicolumn{4}{l}{\textit{SpeechCOMET}} \\
\midrule
\multirow{2}{*}{\textsc{Text}} & COMETKiwi$_{\text{RoBERTa}}^{\text{WMT}}$ & \cellcolor[RGB]{234,246,245}23.1 & \cellcolor[RGB]{231,239,247}-11.1 \\
 & COMETKiwi$_{\text{RoBERTa}}^{\text{IWSLT}}$ & \cellcolor[RGB]{234,246,246}22.8 & \cellcolor[RGB]{229,238,246}-12.0 \\
\multirow{2}{*}{\textsc{Speech}} & SpeechCOMET\textsubscript{SONAR} & \cellcolor[RGB]{239,248,248}20.1 & \cellcolor[RGB]{253,253,254}-3.1 \\
 & SpeechCOMET\textsubscript{Whisper} & \cellcolor[RGB]{244,250,250}17.8 & \cellcolor[RGB]{255,255,255}-2.4 \\
\textsc{Sp.+Txt} & SpeechCOMET & \cellcolor[RGB]{228,243,243}25.8 & \cellcolor[RGB]{213,228,241}-17.8 \\
\textsc{Sp.+Txt} & SpeechCOMET$^{\text{XL}}$ & \cellcolor[RGB]{212,236,236}34.4 & \cellcolor[RGB]{180,206,231}-30.2 \\
\midrule
\multicolumn{4}{l}{\textit{SpeechLLM}} \\
\midrule
\multirow{2}{*}{\textsc{Text}} & SpeechLLM & \cellcolor[RGB]{201,232,231}40.4 & \cellcolor[RGB]{183,208,232}-29.2 \\
 & \quad+FT & \cellcolor[RGB]{188,226,226}47.1 & \cellcolor[RGB]{157,191,223}-38.6 \\
\multirow{2}{*}{\textsc{Speech}} & SpeechLLM & \cellcolor[RGB]{217,238,238}32.0 & \cellcolor[RGB]{186,209,232}-28.1 \\
 & \quad+FT & \cellcolor[RGB]{202,232,231}40.1 & \cellcolor[RGB]{193,214,235}-25.5 \\
\multirow{2}{*}{\textsc{Sp.+Txt}} & SpeechLLM & \cellcolor[RGB]{205,233,233}38.1 & \cellcolor[RGB]{161,193,224}-37.4 \\
 & \quad+FT & \cellcolor[RGB]{183,224,223}49.9 & \cellcolor[RGB]{156,190,223}-39.2 \\
\bottomrule
\end{tabular}
\caption{Effect of replacing the source input with a randomly mismatched one. Scores are averaged over de and zh. $\Delta$ = random$-$real: large negative values indicate the model uses the source signal.}\label{tab:random_src}
\vspace{-2mm}
\end{table}

In contrast, SpeechCOMET\textsubscript{SONAR} and SpeechCOMET\textsubscript{Whisper} drop by only $-3.1$ and $-2.4$ pp respectively, indicating that these models largely ignore the speech input. On the other hand, for the speech+text SpeechCOMET models, the drop with random source ($-17.8$ and $-30.2$ pp) is substantial, but shuffling only the text reproduces it almost entirely while shuffling only the speech has a negligible effect (\cref{tab:source_ablation} in \cref{app:source_ablation}), confirming these models condition on text.

Manual inspection of 50 examples where speech shuffling causes a drop reveals segmentation errors where the speech segment boundaries do not align with the translation (e.g., audio \textit{``Right.''} paired with MT \textit{``Ansätze symmetrisch, richtig?} {\small EN: The approaches are symmetric, right?}''), suggesting that the small residual speech sensitivity reflects misalignments rather than genuine speech conditioning.

Finally, SpeechLLM shows large drops across all modalities ($-25$ to $-39$ pp), including speech-only, confirming that these models do condition on the source speech, yet the probing results show that speech-specific features go unused.

\section{Discussion and Recommendations}\label{sec:discussion}
\paragraph{Three barriers to speech-specific evaluation.}

Our results point to three interacting barriers to speech-specific QE. 
\textit{First}, source encoders discard relevant information: SONAR suppresses acoustic attributes by design, 
while Whisper representations are weak on linguistic features. 
\textit{Second}, even when features are present, models do not use them, 
suggesting QE training fails to ground predictions in the source. 
\textit{Third}, QE fine-tuning does not help, likely due to standard QE corpora lacking examples where speech-specific variation impacts quality. 

\paragraph{Practical guidance for practitioners.}
For practitioners evaluating speech translation quality, we recommend SpeechLLM (+FT) when computational cost is not a constraint, as it achieves the strongest segment-level correlation across all modalities. When a lighter model is needed the off-the-shelf COMETKiwi remains the strongest text-based option. However, none of the evaluated models should be used to assess speech-specific translation quality, such as gender agreement or prosodic appropriateness: all models score at chance on this.

\section{Conclusion}

In this work, we evaluate speech translation quality estimation across regression-based and LLM-based approaches and find that, despite strong overall correlation with human judgements, no current metric is sensitive to speech-specific phenomena such as gender agreement and prosody. Addressing this requires dedicated training data that explicitly targets such phenomena and encoders that capture richer acoustic representations than those induced by the quality estimation objective alone. We hope that our work can serve as a stepping stone for future research on speech-aware quality estimation and encourage the development of benchmarks and models that treat speech translation evaluation as a genuinely multimodal problem.

\section{Limitations}
Correlation results are based on a single test set (IWSLT 2026 Metrics dev set), as no other human-annotated speech translation evaluation datasets are currently available. Similarly, our evaluation of speech-specific phenomena is limited to gender agreement and prosody, and all datasets use English as the source language, as no comparable resources exist for other phenomena or languages. This highlights the need for broader evaluation resources to ensure that findings generalise across languages and domains. 

Beyond generalisability, closing these gaps also matters for societal impact: assessing speech translation quality incorrectly, particularly for speech-specific phenomena such as gender agreement and prosody, can lead to the deployment of systems that misrepresent speakers or lose meaning-bearing cues. Developing reliable methods to evaluate this correctly is therefore essential, and we see our paper as a contribution to this effort.

\section{Ethical Considerations}
This work uses only existing publicly available datasets and pre-trained models. We foresee no direct ethical concerns arising from this research.

\section*{Acknowledgments}
This work has received funding from the European Union’s Horizon research and innovation programme under grant agreement No 101135798, project Meetween (My Personal AI Mediator for Virtual MEETtings BetWEEN People). The authors gratefully acknowledge the computing time provided on the high-performance computer HoreKa by the National High-Performance Computing Center at KIT (NHR@KIT). This center is jointly supported by the Federal Ministry of Education and Research and the Ministry of Science, Research and the Arts of Baden-Württemberg, as part of the National High-Performance Computing (NHR) joint funding program (https://www.nhr-verein.de/en/our-partners). HoreKa is partly funded by the German Research Foundation (DFG).

\bibliography{custom}

\appendix
\crefalias{section}{appendix}
\crefalias{subsection}{appendix}
\crefalias{subsubsection}{appendix}
\section{Training Details}
\subsection{SpeechCOMET}
\label{app:speechcomet}

\cref{tab:speechcomet-hyperparams} lists the hyperparameters used for all SpeechCOMET models.
All models share the same optimizer and estimator architecture; the Whisper LoRA and SONAR fine-tuning settings apply only to the respective encoder variants.
All models are trained on a single NVIDIA A100-SXM4-40GB GPU; speech models take approximately 30h to train, while text-only models require around 20h.

\begin{table}[ht]
\centering
\resizebox{\linewidth}{!}{%
\begin{tabular}{ll}
\toprule
\multicolumn{2}{l}{\textbf{Training}} \\
\midrule
Optimizer & AdamW \\
Learning rate (estimator) & $1.5 \times 10^{-5}$ \\
Learning rate (encoder) & $1 \times 10^{-6}$ \\
Layerwise LR decay & 0.95 \\
Batch size & 2 \\
Gradient accumulation & 8 \\
Max epochs & 20 \\
Gradient clipping & 1.0 \\
Encoder warm-up & 30\% of epoch 0 frozen \\
Dropout & 0.1 \\
Early stopping patience & 2 (val Kendall $\tau$) \\
\midrule
\multicolumn{2}{l}{\textbf{Architecture}} \\
\midrule
Text encoder pooling & layerwise attention (sparsemax) \\
Aggregation & average pooling \\
Estimator hidden sizes & [2048, 1024] \\
Estimator activation & Tanh \\
\midrule
\multicolumn{2}{l}{\textbf{Whisper LoRA}} \\
\midrule
LoRA rank $r$ & 8 \\
LoRA $\alpha$ & 16 \\
LoRA dropout & 0.0 \\
\midrule
\multicolumn{2}{l}{\textbf{SONAR Fine-tuning}} \\
\midrule
Frozen layers (bottom) & 4 of 24 \\
\bottomrule
\end{tabular}%
}
\caption{Hyperparameters for SpeechCOMET models. The effective batch size is 16. All models are trained on a single NVIDIA A100-SXM4-40GB GPU.}\vspace{-0.4cm}
\label{tab:speechcomet-hyperparams}
\end{table}

\subsection{SpeechLLM Details}\label{app:speechllm}

We evaluate \hfmodel{Qwen/Qwen2.5-Omni-7B} \citep{xu2025qwen25omnitechnicalreport} in
both zero-shot and fine-tuned settings across three input modalities for the QE task: text, audio, and
audio+text. For zero-shot evaluation we assess all three modalities of the unmodified base
model. In addition to the standard QE prompt, we evaluate two task-specific prompt
variants: a gender-aware prompt for MuST-SHE and a prosody-aware prompt for ContraProST.
All prompts are shown in \cref{fig:speechllm_prompts}.

For fine-tuning, we train one model per input modality using LlamaFactory~\citep{zheng2024llamafactory}
with LoRA adapters~\citep{lora-hu-2021}. Fine-tuning and inference are performed on a
single NVIDIA A100-SXM4-40GB GPU. Fine-tuning takes around 13 hours. Detailed hyperparameters are listed in
\cref{tab:qwen-hyperparams}.

\begin{table}[ht]
\centering
\small
\begin{tabular}{ll}
\toprule
\multicolumn{2}{l}{\textbf{Fine-tuning}} \\
\midrule
LoRA rank $r$ & 16 \\
LoRA $\alpha$ & 32 \\
LoRA target & all (attention + MLP) \\
Epochs & 2 \\
Learning rate & $1 \times 10^{-4}$ \\
LR scheduler & cosine \\
Warmup ratio & 0.1 \\
Batch size & 1 \\
Gradient accumulation & 8 \\
Precision & bf16 \\
Max input length & 2048 \\
Early stopping patience & 3 \\
Eval interval & 500 steps \\
\midrule
\multicolumn{2}{l}{\textbf{Inference}} \\
\midrule
Decoding & greedy \\
Max new tokens & 16 \\
Precision & bf16 \\
\bottomrule
\end{tabular}
\caption{Fine-tuning and inference hyperparameters for Qwen2.5-Omni-7B.}
\label{tab:qwen-hyperparams}
\end{table}

\begin{figure*}
    \centering

\begin{promptbox}{Standard QE}
\textbf{System:} You are an evaluator. Given the source and/or audio and a
translation, respond with only a single float score between 0 and 1 indicating
translation quality. Output nothing else.\\[4pt]
\textbf{User (text):} Source: \textnormal{\textit{\{src\}}}\\
Translation: \textnormal{\textit{\{mt\}}}\\[4pt]
\textbf{User (audio):} \textnormal{\textit{[audio waveform]}}\\
Translation: \textnormal{\textit{\{mt\}}}\\[4pt]
\textbf{User (audio+text):} \textnormal{\textit{[audio waveform]}}\\
Source: \textnormal{\textit{\{src\}}}\\
Translation: \textnormal{\textit{\{mt\}}}
\end{promptbox}

\begin{promptbox}{Gender-aware QE (MuST-SHE)}
\textbf{System:} You are a translation quality evaluator specialising in gender
accuracy. Listen carefully to the speaker's voice in the audio to determine their
gender. The speaker's gender must be correctly reflected in the translation through
gendered forms such as adjectives, pronouns, and verb agreement. Score the
translation from 0 to 1: give a high score if the gender is correctly reflected,
a low score if it is wrong or ambiguous. Output only a single float score, nothing else.\\[4pt]
\textbf{User (text):} Source: \textnormal{\textit{\{src\}}}\\
Translation: \textnormal{\textit{\{mt\}}}\\[4pt]
\textbf{User (audio):} \textnormal{\textit{[audio waveform]}}\\
Translation: \textnormal{\textit{\{mt\}}}\\[4pt]
\textbf{User (audio+text):} \textnormal{\textit{[audio waveform]}}\\
Source: \textnormal{\textit{\{src\}}}\\
Translation: \textnormal{\textit{\{mt\}}}
\end{promptbox}

\begin{promptbox}{Prosody-aware QE (ContraProST)}
\textbf{System:} You are a translation quality evaluator specialising in prosodic
meaning. The speaker's prosody in the audio --- including intonation, stress,
rhythm, pauses, and emotional tone --- conveys meaning that must be accurately
reflected in the translation (e.g.\ whether an utterance is a question or
statement, which word is emphasised, the level of politeness or emotion). Score
the translation from 0 to 1 based on how well it captures the prosodic meaning
of the source speech. Output only a single float score, nothing else.\\[4pt]
\textbf{User (text):} Source: \textnormal{\textit{\{src\}}}\\
Translation: \textnormal{\textit{\{mt\}}}\\[4pt]
\textbf{User (audio):} \textnormal{\textit{[audio waveform]}}\\
Translation: \textnormal{\textit{\{mt\}}}\\[4pt]
\textbf{User (audio+text):} \textnormal{\textit{[audio waveform]}}\\
Source: \textnormal{\textit{\{src\}}}\\
Translation: \textnormal{\textit{\{mt\}}}
\end{promptbox}

    \caption{System and user prompts used for Qwen2.5-Omni-7B QE. The system prompt varies across the three settings: standard translation quality, gender accuracy, and prosodic meaning. The user prompt is identical across all three settings and varies only by input modality.}
    \label{fig:speechllm_prompts}
\end{figure*}

\subsection{Training Data}\label{app:data}

\paragraph{IWSLT training set.}
The \hfmodel{maikezu/IWSLT2026-metrics-shared-train-dev} dataset combines three sources of human direct assessment annotations, all paired with source speech: IWSLT 2023~\citep{agrawal-etal-2023-findings}, where evaluators had access to one preceding and one following segment for context; and WMT 2024~\citep{kocmi-etal-2024-findings} and WMT 2025~\citep{kocmi-etal-2025-findings}, both evaluated on human-segmented audio. It covers 17 language pairs: en-ar, en-bho, en-cs, en-de, en-es, en-et, en-hi, en-is, en-it, en-ja, en-mas, en-ru, en-sr, en-uk, en-zh, cs-de, and cs-uk.

\paragraph{WMT augmentation.}
The \hfmodel{zouharvi/WMT-human-all} dataset aggregates human assessments from multiple WMT editions. We exclude the WMT 2024 and WMT 2025 subsets that overlap with the IWSLT training set. The remaining data covers text translation systems with gold sentence segmentation and 49 language pairs: bn-hi, cs-de, cs-en, cs-uk, de-cs, de-en, de-fr, en-ar, en-bho, en-cs, en-de, en-es, en-et, en-ha, en-hi, en-hr, en-is, en-it, en-iu, en-ja, en-kk, en-ko, en-liv, en-mas, en-pl, en-ru, en-sr, en-uk, en-zh, fi-en, fr-de, gu-en, ha-en, he-en, hi-bn, is-en, ja-en, ja-zh, kk-en, liv-en, lt-en, pl-en, ps-en, ru-en, sah-ru, uk-en, xh-zu, zh-en, and zu-xh.

\begin{table*}[t]
\centering
\footnotesize
\setlength{\tabcolsep}{3pt}
\begin{tabular}{ll|rrr|rrr|rrr|rrr}
\toprule
 & & \multicolumn{6}{c|}{\textbf{IWSLT dev}} & \multicolumn{3}{c|}{\textbf{MuST-SHE}} & \multicolumn{3}{c}{\textbf{ContraProST}} \\
 & & \multicolumn{3}{c|}{Segment $\tau_b$ (\%)} & \multicolumn{3}{c|}{System SPA (\%)} & \multicolumn{3}{c|}{PA (\%)} & \multicolumn{3}{c}{PA (\%)} \\
\cmidrule(lr){3-5}\cmidrule(lr){6-8}\cmidrule(lr){9-11}\cmidrule(lr){12-14}
\textbf{Fusion} & \textbf{SONAR} & \textbf{de} & \textbf{zh} & \textbf{avg} & \textbf{de} & \textbf{zh} & \textbf{avg} & \textbf{es} & \textbf{fr} & \textbf{it} & \textbf{de} & \textbf{es} & \textbf{ja} \\
\midrule
avg & frozen & \cellcolor[RGB]{227,243,242}17.0 & \cellcolor[RGB]{214,237,237}24.5 & \cellcolor[RGB]{222,241,240}20.8 & \cellcolor[RGB]{219,239,239}64.5 & \cellcolor[RGB]{252,253,253}67.7 & \cellcolor[RGB]{231,244,244}66.1 & \cellcolor[RGB]{252,221,197}52.8 & \cellcolor[RGB]{252,215,186}53.0 & \cellcolor[RGB]{253,238,226}50.7 & \cellcolor[RGB]{232,168,205}\textbf{50.1} & \cellcolor[RGB]{232,168,205}\textbf{50.1} & \cellcolor[RGB]{255,255,255}\textbf{50.0} \\
\cmidrule(l){1-14}
concat & frozen & \cellcolor[RGB]{255,255,255}12.2 & \cellcolor[RGB]{255,255,255}21.1 & \cellcolor[RGB]{255,255,255}16.6 & \cellcolor[RGB]{255,255,255}49.0 & \cellcolor[RGB]{183,224,223}\textbf{75.7} & \cellcolor[RGB]{255,255,255}62.3 & \cellcolor[RGB]{253,226,205}52.4 & \cellcolor[RGB]{255,255,255}48.6 & \cellcolor[RGB]{252,224,201}51.3 & \cellcolor[RGB]{255,255,255}50.0 & \cellcolor[RGB]{255,255,255}50.0 & \cellcolor[RGB]{255,255,255}\textbf{50.0} \\
\cmidrule(l){1-14}
\multirow{3}{*}{sum} & frozen & \cellcolor[RGB]{204,233,233}20.9 & \cellcolor[RGB]{212,236,236}24.7 & \cellcolor[RGB]{207,234,234}22.8 & \cellcolor[RGB]{206,234,233}70.2 & \cellcolor[RGB]{255,255,255}67.4 & \cellcolor[RGB]{214,237,237}68.8 & \cellcolor[RGB]{251,197,156}54.8 & \cellcolor[RGB]{250,183,132}\textbf{55.4} & \cellcolor[RGB]{250,183,132}\textbf{53.0} & \cellcolor[RGB]{255,255,255}50.0 & \cellcolor[RGB]{255,255,255}50.0 & \cellcolor[RGB]{255,255,255}\textbf{50.0} \\
 & \quad WMT text pretrain $\to$ FT (frozen) & \cellcolor[RGB]{183,224,223}\textbf{24.5} & \cellcolor[RGB]{183,224,223}\textbf{27.1} & \cellcolor[RGB]{183,224,223}\textbf{25.8} & \cellcolor[RGB]{184,224,224}79.7 & \cellcolor[RGB]{255,255,255}67.4 & \cellcolor[RGB]{185,225,224}73.5 & \cellcolor[RGB]{250,188,140}55.6 & \cellcolor[RGB]{252,220,195}52.6 & \cellcolor[RGB]{251,197,156}52.4 & \cellcolor[RGB]{255,255,255}50.0 & \cellcolor[RGB]{255,255,255}49.9 & \cellcolor[RGB]{255,255,255}\textbf{50.0} \\
 & \quad WMT text pretrain $\to$ FT (unfrozen) & \cellcolor[RGB]{189,227,226}23.4 & \cellcolor[RGB]{183,224,223}\textbf{27.1} & \cellcolor[RGB]{187,226,225}25.2 & \cellcolor[RGB]{183,224,223}\textbf{80.2} & \cellcolor[RGB]{255,255,255}67.4 & \cellcolor[RGB]{183,224,223}\textbf{73.8} & \cellcolor[RGB]{250,183,132}\textbf{56.0} & \cellcolor[RGB]{251,196,154}54.4 & \cellcolor[RGB]{250,183,132}\textbf{53.0} & \cellcolor[RGB]{255,255,255}50.0 & \cellcolor[RGB]{255,255,255}50.0 & \cellcolor[RGB]{255,255,255}\textbf{50.0} \\
\bottomrule
\end{tabular}
\caption{Ablation of embedding fusion strategies for \textsc{Text + Speech} SpeechCOMET with SONAR encoder. \emph{Fusion} controls how speech and text embeddings are combined; \emph{SONAR} indicates whether the encoder is frozen or unfrozen during fine-tuning. Text init initialises from the WMT-pretrained text SpeechCOMET checkpoint.}\label{tab:orkney_ablation}
\end{table*}

\begin{table*}[t]
\centering
\footnotesize
\setlength{\tabcolsep}{4pt}
\begin{tabular}{ll|ccc|ccc}
\toprule
 & & \multicolumn{3}{c|}{\textbf{MuST-SHE PA (\%)}} & \multicolumn{3}{c}{\textbf{ContraProST PA (\%)}} \\
\cmidrule(lr){3-5}\cmidrule(lr){6-8}
\textbf{Type} & \textbf{Model} & \textbf{es} & \textbf{fr} & \textbf{it} & \textbf{de} & \textbf{es} & \textbf{ja} \\
\midrule
\multirow{3}{*}{\textsc{Speech}} & SpeechLLM & \cellcolor[RGB]{255,255,255}31.9 & \cellcolor[RGB]{255,255,255}28.3 & \cellcolor[RGB]{255,255,255}32.2 & \cellcolor[RGB]{255,255,255}\textbf{31.6} & \cellcolor[RGB]{255,255,255}\textbf{28.4} & \cellcolor[RGB]{255,255,255}\textbf{31.6} \\
 & \quad speech prompt & \cellcolor[RGB]{255,255,255}38.0 & \cellcolor[RGB]{255,255,255}32.3 & \cellcolor[RGB]{255,255,255}32.4 & \cellcolor[RGB]{255,255,255}15.3 & \cellcolor[RGB]{255,255,255}19.4 & \cellcolor[RGB]{255,255,255}16.8 \\
 & \quad+FT & \cellcolor[RGB]{255,255,255}35.9 & \cellcolor[RGB]{255,255,255}15.0 & \cellcolor[RGB]{255,255,255}20.7 & \cellcolor[RGB]{255,255,255}20.8 & \cellcolor[RGB]{255,255,255}22.1 & \cellcolor[RGB]{255,255,255}15.3 \\
\midrule
\multirow{3}{*}{\textsc{Text}} & SpeechLLM & \cellcolor[RGB]{255,255,255}29.4 & \cellcolor[RGB]{255,255,255}25.9 & \cellcolor[RGB]{255,255,255}23.6 & \cellcolor[RGB]{255,255,255}23.8 & \cellcolor[RGB]{255,255,255}18.6 & \cellcolor[RGB]{255,255,255}29.6 \\
 & \quad speech prompt & \cellcolor[RGB]{255,255,255}23.7 & \cellcolor[RGB]{255,255,255}17.3 & \cellcolor[RGB]{255,255,255}16.2 & \cellcolor[RGB]{255,255,255}18.9 & \cellcolor[RGB]{255,255,255}18.5 & \cellcolor[RGB]{255,255,255}15.7 \\
 & \quad+FT & \cellcolor[RGB]{255,255,255}14.0 & \cellcolor[RGB]{255,255,255}13.8 & \cellcolor[RGB]{255,255,255}15.4 & \cellcolor[RGB]{255,255,255}14.1 & \cellcolor[RGB]{255,255,255}14.1 & \cellcolor[RGB]{255,255,255}25.0 \\
\midrule
\multirow{3}{*}{\textsc{Sp.+Txt}} & SpeechLLM & \cellcolor[RGB]{255,255,255}38.0 & \cellcolor[RGB]{255,255,255}30.9 & \cellcolor[RGB]{255,255,255}34.5 & \cellcolor[RGB]{255,255,255}30.5 & \cellcolor[RGB]{255,255,255}26.4 & \cellcolor[RGB]{255,255,255}31.3 \\
 & \quad speech prompt & \cellcolor[RGB]{255,255,255}\textbf{43.1} & \cellcolor[RGB]{255,255,255}\textbf{37.1} & \cellcolor[RGB]{255,255,255}\textbf{37.8} & \cellcolor[RGB]{255,255,255}22.8 & \cellcolor[RGB]{255,255,255}26.8 & \cellcolor[RGB]{255,255,255}21.4 \\
 & \quad+FT & \cellcolor[RGB]{255,255,255}8.2 & \cellcolor[RGB]{255,255,255}6.3 & \cellcolor[RGB]{255,255,255}6.8 & \cellcolor[RGB]{255,255,255}18.0 & \cellcolor[RGB]{255,255,255}9.3 & \cellcolor[RGB]{255,255,255}19.7 \\
\bottomrule
\end{tabular}
\caption{Effect of speech-aware prompting on SpeechLLM across input modalities. The speech prompt instructs the model to consider paralinguistic cues; +FT denotes fine-tuning on IWSLT.}\label{tab:speechllm_prompt}
\end{table*}

\paragraph{TTS synthesis.}
Since WMT provides only source text, we synthesise speech using \hfmodel{hexgrad/Kokoro-82M} for eight languages (English, Japanese, Chinese, Spanish, French, Hindi, Italian, Portuguese) and \hfmodel{facebook/MMS-TTS} as a fallback for the remaining languages. Voice selection is deterministic per source segment. Identical source texts receive the same audio. The resulting dataset contains 1{,}114 hours.

\section{Detailed Results}\label{app:results}
\subsection{SpeechCOMET Fusion Strategies}\label{app:harris_fusion}
\begin{table*}[t]
\centering
\footnotesize
\setlength{\tabcolsep}{3pt}
\begin{tabular}{l|rrr|rrr|rrr|rrr}
\toprule
 & \multicolumn{6}{c|}{\textbf{IWSLT dev}} & \multicolumn{3}{c|}{\textbf{MuST-SHE}} & \multicolumn{3}{c}{\textbf{ContraProST}} \\
 & \multicolumn{3}{c|}{Segment $\tau_b$ (\%)} & \multicolumn{3}{c|}{System SPA (\%)} & \multicolumn{3}{c|}{PA (\%)} & \multicolumn{3}{c}{PA (\%)} \\
\cmidrule(lr){2-4}\cmidrule(lr){5-7}\cmidrule(lr){8-10}\cmidrule(lr){11-13}
\textbf{Architecture} & \textbf{de} & \textbf{zh} & \textbf{avg} & \textbf{de} & \textbf{zh} & \textbf{avg} & \textbf{es} & \textbf{fr} & \textbf{it} & \textbf{de} & \textbf{es} & \textbf{ja} \\
\midrule
four-way & \cellcolor[RGB]{183,224,223}\textbf{14.6} & \cellcolor[RGB]{183,224,223}\textbf{22.5} & \cellcolor[RGB]{183,224,223}\textbf{18.6} & \cellcolor[RGB]{183,224,223}\textbf{37.2} & \cellcolor[RGB]{211,236,235}80.4 & \cellcolor[RGB]{183,224,223}\textbf{58.8} & \cellcolor[RGB]{250,183,132}\textbf{56.0} & \cellcolor[RGB]{250,183,132}\textbf{51.9} & \cellcolor[RGB]{252,221,197}51.3 & \cellcolor[RGB]{255,255,255}50.0 & \cellcolor[RGB]{255,255,255}\textbf{50.0} & \cellcolor[RGB]{255,255,255}\textbf{50.0} \\
\midrule
additive & \cellcolor[RGB]{190,227,227}13.5 & \cellcolor[RGB]{184,224,224}22.3 & \cellcolor[RGB]{187,225,225}17.9 & \cellcolor[RGB]{191,227,227}35.3 & \cellcolor[RGB]{255,255,255}67.4 & \cellcolor[RGB]{255,255,255}51.4 & \cellcolor[RGB]{251,200,160}54.6 & \cellcolor[RGB]{255,255,255}49.7 & \cellcolor[RGB]{251,198,158}52.2 & \cellcolor[RGB]{255,255,255}50.0 & \cellcolor[RGB]{255,255,255}\textbf{50.0} & \cellcolor[RGB]{255,255,255}\textbf{50.0} \\
\midrule
joint encoding & \cellcolor[RGB]{255,255,255}3.9 & \cellcolor[RGB]{255,255,255}6.9 & \cellcolor[RGB]{255,255,255}5.4 & \cellcolor[RGB]{255,255,255}19.9 & \cellcolor[RGB]{183,224,223}\textbf{88.6} & \cellcolor[RGB]{227,243,242}54.2 & \cellcolor[RGB]{253,235,222}51.6 & \cellcolor[RGB]{254,243,235}50.3 & \cellcolor[RGB]{250,183,132}\textbf{52.8} & \cellcolor[RGB]{232,168,205}\textbf{50.6} & \cellcolor[RGB]{255,255,255}49.2 & \cellcolor[RGB]{255,255,255}\textbf{50.0} \\
\bottomrule
\end{tabular}
\caption{Ablation of audio--text integration strategies for speech QE. \emph{Four-way}: standard COMET estimator input $[\mathbf{h}_t; \mathbf{s}_a; \mathbf{h}_t \odot \mathbf{s}_a; |\mathbf{h}_t - \mathbf{s}_a|]$, with SONAR audio replacing the source text embedding (used in the main model). \emph{Additive}: SONAR audio embedding summed with the unified CLS MT embedding, following the COMETKiwi unified-metric architecture. \emph{Joint encoding}: SONAR audio embedding injected as a virtual token into XLM-R self-attention. All models are speech-only (SONAR audio input) fine-tuned on IWSLT.}
\label{tab:harris_ablation}
\end{table*}

\begin{table*}[t]
\centering
\footnotesize
\setlength{\tabcolsep}{3pt}
\begin{tabular}{p{3cm}l|rrr|rrr|rrr|rrr}
\toprule
 & & \multicolumn{6}{c|}{\textbf{IWSLT dev}} & \multicolumn{3}{c|}{\textbf{MuST-SHE}} & \multicolumn{3}{c}{\textbf{ContraProST}} \\
 & & \multicolumn{3}{c|}{Segment $\tau_b$ (\%)} & \multicolumn{3}{c|}{System SPA (\%)} & \multicolumn{3}{c|}{PA (\%)} & \multicolumn{3}{c}{PA (\%)} \\
\cmidrule(lr){3-5}\cmidrule(lr){6-8}\cmidrule(lr){9-11}\cmidrule(lr){12-14}
\textbf{Model} & \textbf{Variant} & \textbf{de} & \textbf{zh} & \textbf{avg} & \textbf{de} & \textbf{zh} & \textbf{avg} & \textbf{es} & \textbf{fr} & \textbf{it} & \textbf{de} & \textbf{es} & \textbf{ja} \\
\midrule
\multicolumn{14}{l}{\textit{SONAR encoder}} \\
\midrule
\multirow{2}{*}{\parbox[t]{3cm}{\raggedright IWSLT FT}} & frozen & \cellcolor[RGB]{201,232,231}14.6 & \cellcolor[RGB]{186,225,225}22.5 & \cellcolor[RGB]{196,230,229}18.5 & \cellcolor[RGB]{219,239,239}37.2 & \cellcolor[RGB]{183,224,223}\textbf{80.4} & \cellcolor[RGB]{195,229,228}58.8 & \cellcolor[RGB]{250,188,139}56.0 & \cellcolor[RGB]{252,221,196}51.9 & \cellcolor[RGB]{253,234,220}51.3 & \cellcolor[RGB]{255,255,255}50.0 & \cellcolor[RGB]{255,255,255}50.0 & \cellcolor[RGB]{255,255,255}50.0 \\
 & unfrozen & \cellcolor[RGB]{183,224,223}\textbf{17.3} & \cellcolor[RGB]{183,224,223}\textbf{22.9} & \cellcolor[RGB]{183,224,223}\textbf{20.1} & \cellcolor[RGB]{183,224,223}\textbf{51.0} & \cellcolor[RGB]{216,238,238}70.9 & \cellcolor[RGB]{183,224,223}\textbf{61.0} & \cellcolor[RGB]{252,220,195}53.1 & \cellcolor[RGB]{253,238,227}50.9 & \cellcolor[RGB]{254,241,230}50.9 & \cellcolor[RGB]{255,255,255}50.0 & \cellcolor[RGB]{255,255,255}50.0 & \cellcolor[RGB]{255,255,255}50.0 \\
\cmidrule(l){1-14}
\multirow{2}{*}{\parbox[t]{3cm}{\raggedright WMT text pretrain $\to$ IWSLT FT}} & frozen & \cellcolor[RGB]{195,229,228}15.6 & \cellcolor[RGB]{209,235,235}19.3 & \cellcolor[RGB]{206,234,233}17.4 & \cellcolor[RGB]{209,235,235}41.1 & \cellcolor[RGB]{225,242,242}68.3 & \cellcolor[RGB]{216,238,238}54.7 & \cellcolor[RGB]{252,220,195}53.1 & \cellcolor[RGB]{251,201,162}53.0 & \cellcolor[RGB]{254,253,252}50.1 & \cellcolor[RGB]{232,168,205}\textbf{50.1} & \cellcolor[RGB]{255,255,255}50.0 & \cellcolor[RGB]{255,255,255}50.0 \\
 & unfrozen & \cellcolor[RGB]{187,226,225}16.7 & \cellcolor[RGB]{196,229,229}21.2 & \cellcolor[RGB]{193,228,228}18.9 & \cellcolor[RGB]{196,229,229}46.2 & \cellcolor[RGB]{207,234,234}73.6 & \cellcolor[RGB]{189,226,226}59.9 & \cellcolor[RGB]{252,215,187}53.5 & \cellcolor[RGB]{252,213,184}52.3 & \cellcolor[RGB]{254,247,241}50.5 & \cellcolor[RGB]{255,255,255}50.0 & \cellcolor[RGB]{255,255,255}50.0 & \cellcolor[RGB]{255,255,255}50.0 \\
\cmidrule(l){1-14}
\multirow{2}{*}{\parbox[t]{3cm}{\raggedright WMT TTS pretrain (frozen)}} & pretrain only & \cellcolor[RGB]{227,243,242}10.7 & \cellcolor[RGB]{229,244,243}16.6 & \cellcolor[RGB]{237,247,247}13.7 & \cellcolor[RGB]{255,255,255}23.6 & \cellcolor[RGB]{214,237,237}71.5 & \cellcolor[RGB]{255,255,255}47.6 & \cellcolor[RGB]{252,213,183}53.7 & \cellcolor[RGB]{252,217,190}52.1 & \cellcolor[RGB]{250,183,132}\textbf{54.6} & \cellcolor[RGB]{255,255,255}49.9 & \cellcolor[RGB]{255,255,255}50.0 & \cellcolor[RGB]{255,255,255}50.0 \\
 & \quad $\to$ IWSLT FT (frozen) & \cellcolor[RGB]{207,234,234}13.8 & \cellcolor[RGB]{213,237,236}18.8 & \cellcolor[RGB]{215,238,237}16.3 & \cellcolor[RGB]{231,244,244}32.7 & \cellcolor[RGB]{227,243,243}67.7 & \cellcolor[RGB]{241,249,248}50.2 & \cellcolor[RGB]{252,213,183}53.7 & \cellcolor[RGB]{252,213,184}52.3 & \cellcolor[RGB]{250,186,137}54.4 & \cellcolor[RGB]{255,255,255}50.0 & \cellcolor[RGB]{255,255,255}49.9 & \cellcolor[RGB]{255,255,255}50.0 \\
\cmidrule(l){2-14}
\multirow{3}{*}{\parbox[t]{3cm}{\raggedright WMT TTS pretrain (unfrozen)}} & pretrain only & \cellcolor[RGB]{221,240,240}11.6 & \cellcolor[RGB]{238,247,247}15.4 & \cellcolor[RGB]{239,248,248}13.5 & \cellcolor[RGB]{248,252,251}26.3 & \cellcolor[RGB]{212,236,236}72.1 & \cellcolor[RGB]{246,251,251}49.2 & \cellcolor[RGB]{254,244,237}50.9 & \cellcolor[RGB]{253,238,227}50.9 & \cellcolor[RGB]{251,202,164}53.4 & \cellcolor[RGB]{255,255,255}50.0 & \cellcolor[RGB]{232,168,205}\textbf{50.2} & \cellcolor[RGB]{255,255,255}49.7 \\
 & \quad $\to$ IWSLT FT (frozen) & \cellcolor[RGB]{198,230,230}15.1 & \cellcolor[RGB]{204,233,232}20.0 & \cellcolor[RGB]{205,233,233}17.6 & \cellcolor[RGB]{231,244,244}32.7 & \cellcolor[RGB]{227,243,242}67.9 & \cellcolor[RGB]{240,248,248}50.3 & \cellcolor[RGB]{250,183,132}\textbf{56.4} & \cellcolor[RGB]{252,224,202}51.7 & \cellcolor[RGB]{251,208,174}53.0 & \cellcolor[RGB]{255,255,255}50.0 & \cellcolor[RGB]{255,255,255}50.0 & \cellcolor[RGB]{247,226,238}50.1 \\
 & \quad $\to$ IWSLT FT (unfrozen) & \cellcolor[RGB]{194,229,228}15.7 & \cellcolor[RGB]{191,227,227}21.8 & \cellcolor[RGB]{195,229,228}18.7 & \cellcolor[RGB]{218,239,239}37.7 & \cellcolor[RGB]{224,242,241}68.6 & \cellcolor[RGB]{225,242,242}53.1 & \cellcolor[RGB]{251,209,176}54.1 & \cellcolor[RGB]{250,183,132}\textbf{54.0} & \cellcolor[RGB]{251,205,169}53.2 & \cellcolor[RGB]{255,255,255}49.9 & \cellcolor[RGB]{255,255,255}50.0 & \cellcolor[RGB]{232,168,205}\textbf{50.3} \\
\midrule
\multicolumn{14}{l}{\textit{Whisper encoder}} \\
\midrule
\multirow{2}{*}{\parbox[t]{3cm}{\raggedright IWSLT FT (avg pool)}} & frozen & \cellcolor[RGB]{228,243,243}10.6 & \cellcolor[RGB]{214,237,237}18.6 & \cellcolor[RGB]{230,244,244}14.6 & \cellcolor[RGB]{213,237,236}39.6 & \cellcolor[RGB]{228,243,243}67.6 & \cellcolor[RGB]{222,241,240}53.6 & \cellcolor[RGB]{254,249,245}50.5 & \cellcolor[RGB]{255,255,255}46.9 & \cellcolor[RGB]{255,255,255}49.1 & \cellcolor[RGB]{255,255,255}50.0 & \cellcolor[RGB]{255,255,255}49.9 & \cellcolor[RGB]{255,255,255}50.0 \\
 & LoRA & \cellcolor[RGB]{247,251,251}7.7 & \cellcolor[RGB]{230,244,244}16.5 & \cellcolor[RGB]{251,253,253}12.1 & \cellcolor[RGB]{224,242,241}35.2 & \cellcolor[RGB]{228,243,243}67.4 & \cellcolor[RGB]{235,246,246}51.3 & \cellcolor[RGB]{251,209,176}54.1 & \cellcolor[RGB]{253,226,205}51.6 & \cellcolor[RGB]{255,255,255}49.1 & \cellcolor[RGB]{232,168,205}\textbf{50.1} & \cellcolor[RGB]{255,255,255}50.0 & \cellcolor[RGB]{255,255,255}49.9 \\
\multirow{2}{*}{\parbox[t]{3cm}{\raggedright IWSLT FT (attn pool)}} & frozen & \cellcolor[RGB]{255,255,255}6.6 & \cellcolor[RGB]{228,243,243}16.7 & \cellcolor[RGB]{255,255,255}11.6 & \cellcolor[RGB]{200,231,231}44.4 & \cellcolor[RGB]{228,243,243}67.4 & \cellcolor[RGB]{210,235,235}55.9 & \cellcolor[RGB]{250,190,143}\textbf{55.8} & \cellcolor[RGB]{253,238,227}50.9 & \cellcolor[RGB]{253,225,204}51.9 & \cellcolor[RGB]{232,168,205}\textbf{50.1} & \cellcolor[RGB]{255,255,255}49.9 & \cellcolor[RGB]{255,255,255}50.0 \\
 & LoRA & \cellcolor[RGB]{245,251,251}7.9 & \cellcolor[RGB]{208,235,234}\textbf{19.4} & \cellcolor[RGB]{237,247,247}13.7 & \cellcolor[RGB]{205,233,233}42.5 & \cellcolor[RGB]{228,243,243}67.4 & \cellcolor[RGB]{215,238,237}54.9 & \cellcolor[RGB]{250,192,147}55.6 & \cellcolor[RGB]{252,221,196}51.9 & \cellcolor[RGB]{251,202,164}\textbf{53.4} & \cellcolor[RGB]{255,255,255}50.0 & \cellcolor[RGB]{243,211,230}50.1 & \cellcolor[RGB]{255,255,255}50.0 \\
\cmidrule(l){1-14}
\multirow{3}{*}{\parbox[t]{3cm}{\raggedright WMT TTS pretrain (LoRA)}} & pretrain only & \cellcolor[RGB]{219,239,239}11.9 & \cellcolor[RGB]{255,255,255}13.1 & \cellcolor[RGB]{247,251,251}12.5 & \cellcolor[RGB]{196,229,229}\textbf{46.1} & \cellcolor[RGB]{255,255,255}59.9 & \cellcolor[RGB]{226,242,242}53.0 & \cellcolor[RGB]{255,255,255}49.5 & \cellcolor[RGB]{251,196,153}\textbf{53.3} & \cellcolor[RGB]{255,255,255}48.1 & \cellcolor[RGB]{255,255,255}49.9 & \cellcolor[RGB]{232,168,205}\textbf{50.2} & \cellcolor[RGB]{255,255,255}49.9 \\
 & \quad $\to$ IWSLT FT (frozen) & \cellcolor[RGB]{186,225,225}\textbf{16.8} & \cellcolor[RGB]{213,237,237}18.7 & \cellcolor[RGB]{203,232,232}\textbf{17.8} & \cellcolor[RGB]{199,231,230}44.9 & \cellcolor[RGB]{225,242,241}\textbf{68.5} & \cellcolor[RGB]{206,234,233}\textbf{56.7} & \cellcolor[RGB]{253,239,228}51.4 & \cellcolor[RGB]{254,242,233}50.7 & \cellcolor[RGB]{252,211,180}52.8 & \cellcolor[RGB]{255,255,255}50.0 & \cellcolor[RGB]{255,255,255}50.0 & \cellcolor[RGB]{247,226,238}\textbf{50.1} \\
 & \quad $\to$ IWSLT FT (LoRA) & \cellcolor[RGB]{191,227,227}16.1 & \cellcolor[RGB]{209,235,234}19.4 & \cellcolor[RGB]{203,232,232}17.7 & \cellcolor[RGB]{208,234,234}41.6 & \cellcolor[RGB]{225,242,241}\textbf{68.5} & \cellcolor[RGB]{215,237,237}55.1 & \cellcolor[RGB]{254,251,249}50.3 & \cellcolor[RGB]{254,246,239}50.5 & \cellcolor[RGB]{252,217,190}52.4 & \cellcolor[RGB]{255,255,255}49.9 & \cellcolor[RGB]{255,255,255}49.9 & \cellcolor[RGB]{255,255,255}50.0 \\
\bottomrule
\end{tabular}
\caption{Ablation of pretraining strategies for SpeechCOMET. $\to$ denotes sequential training stages. TTS pretraining uses WMT data with text-to-speech synthesis.}\label{tab:tts_ablation}
\end{table*}

\cref{tab:harris_ablation} compares three strategies for fusing the SONAR audio embedding $\mathbf{s}_a$ with the translation representation $\mathbf{h}_t$ before the regression head. All variants use XLM-RoBERTa-base with SONAR audio input, fine-tuned on IWSLT.

\paragraph{Four-way interaction.}
Following the original COMET estimator design, the two representations are combined as $[\mathbf{h}_t;\, \mathbf{s}_a;\, \mathbf{h}_t \odot \mathbf{s}_a;\, |\mathbf{h}_t - \mathbf{s}_a|]$, with SONAR audio replacing the text source embedding. This is the architecture used for SpeechCOMET in the main experiments and achieves the best segment-level correlation among the variants.

\paragraph{Additive.}
Following the COMETKiwi unified-metric design, the audio embedding is added to the MT CLS representation before the regression head, collapsing both modalities into a  single vector without explicit interaction terms.

\paragraph{Joint encoding.}
The SONAR audio embedding is appended as a virtual token to the MT token embeddings before the XLM-R transformer layers, enabling full cross-attention between the audio  representation and every MT subword. The resulting CLS representation is fed directly to the regression head.
Despite the richer interaction, this variant yields substantially lower segment-level correlation, suggesting that injecting a fixed-size audio embedding as a single token is difficult for the transformer to exploit effectively.

\subsection{Training Strategies for SpeechCOMET}\label{app:training_strat}
\cref{tab:tts_ablation} reports the full ablation over encoder fine-tuning and TTS pretraining strategies for SONAR and Whisper speech encoders.

\paragraph{SONAR.}
Unfreezing the SONAR encoder consistently improves segment-level correlation over the frozen variant (20.1 vs.\ 18.5 $\tau_b$ avg for direct IWSLT fine-tuning). Neither WMT text pretraining nor TTS-synthesised WMT pretraining improves over direct IWSLT fine-tuning, with most pretraining variants performing below the unfrozen IWSLT baseline. This suggests that SONAR's fixed-size sentence-level representations already provide a good initialisation for quality estimation, and additional pretraining on out-of-domain data adds little.

\paragraph{Whisper.}
Without pretraining, Whisper performs substantially below SONAR regardless of pooling strategy or adapter configuration (best: 14.6 $\tau_b$ avg). TTS pretraining on WMT data followed by frozen IWSLT fine-tuning is the strongest configuration (17.8 $\tau_b$, 56.7 SPA avg), confirming that Whisper benefits from exposure to additional speech data before task-specific training. MuST-SHE and ContraProST scores remain near chance across all variants for both encoders.

\subsection{Fusion Strategies for SpeechCOMET (Speech+Text)}\label{app:fusion_speech_text}

\cref{tab:orkney_ablation} ablates embedding fusion strategy and SONAR encoder state for SpeechCOMET models combining speech and text source representations.

\paragraph{Fusion strategy.}

Sum fusion outperforms both average and concatenation on segment-level correlation. Concatenation performs worst (16.6 $\tau_b$ avg). Average fusion (20.8 $\tau_b$) falls below sum (22.8 $\tau_b$).

\paragraph{Initialisation from text checkpoint.}
Initialising the multimodal model from the WMT-pretrained text SpeechCOMET checkpoint before IWSLT fine-tuning yields a substantial gain over training from scratch (25.8 vs.\ 22.8 $\tau_b$ avg). Freezing vs.\ unfreezing SONAR during this stage makes little difference. MuST-SHE scores remain near chance across all variants and ContraProST shows no sensitivity.

\begin{table*}[ht!]
\centering
\footnotesize
\begin{tabular}{lccc}
\toprule
& \multicolumn{3}{c}{Segment $\tau_b$ (\%)} \\
\cmidrule(lr){2-4}
\textbf{Model} & \textbf{de} & \textbf{zh} & \textbf{avg} \\
\midrule
COMET-PARTIAL & 11.2$\pm$3.5 & 12.7$\pm$6.8 & 11.9$\pm$3.7 \\
COMETKiwi & 32.8$\pm$3.2$^\dagger$ & 36.4$\pm$6.1 & 34.6$\pm$3.4$^\dagger$ \\
SpeechQE & 26.9$\pm$3.3 & 32.7$\pm$6.7 & 29.8$\pm$3.7 \\
BLASER & 22.0$\pm$3.2 & 26.8$\pm$6.3 & 24.4$\pm$3.5 \\
\midrule
COMETKiwi$_{\text{RoBERTa}}^{\text{WMT}}$ & 20.3$\pm$3.4 & 25.9$\pm$6.4 & 23.1$\pm$3.7 \\
COMETKiwi$_{\text{RoBERTa}}^{\text{IWSLT}}$ & 20.2$\pm$3.2 & 25.3$\pm$6.8 & 22.8$\pm$3.5 \\
SpeechCOMET\textsubscript{SONAR} & 17.3$\pm$3.4 & 22.9$\pm$6.4 & 20.1$\pm$3.9 \\
SpeechCOMET\textsubscript{Whisper} & 16.8$\pm$3.4 & 18.7$\pm$6.2 & 17.8$\pm$3.7 \\
SpeechCOMET & 24.5$\pm$3.2 & 27.1$\pm$6.4 & 25.8$\pm$3.6 \\
SpeechCOMET$^{\text{XL}}$ & 32.7$\pm$3.1$^\dagger$ & 36.0$\pm$6.2 & 34.4$\pm$3.7 \\
\midrule
SpeechLLM (\textsc{Text}) & 33.2$\pm$3.4$^\dagger$ & 47.5$\pm$6.5$^\dagger$ & 40.4$\pm$3.7$^\dagger$ \\
\quad +FT & 39.3$\pm$3.6$^\dagger$ & 55.0$\pm$7.0$^\dagger$ & 47.1$\pm$4.1$^\dagger$ \\
SpeechLLM (\textsc{Speech}) & 26.5$\pm$3.6 & 37.5$\pm$7.0 & 32.0$\pm$3.8 \\
\quad +FT & 33.8$\pm$4.4 & 46.5$\pm$7.4$^\dagger$ & 40.1$\pm$4.4 \\
SpeechLLM (\textsc{Sp.+Txt}) & 32.2$\pm$3.1$^\dagger$ & 44.1$\pm$6.8$^\dagger$ & 38.1$\pm$3.8$^\dagger$ \\
\quad +FT & \textbf{39.7$\pm$4.0} & \textbf{60.1$\pm$7.1} & \textbf{49.9$\pm$4.1} \\
\bottomrule
\end{tabular}
\caption{Full IWSLT dev results across all models. Bold marks the column's single best model; $^\dagger$ indicates not significantly different from the best (paired bootstrap, $p\geq0.05$). Plain values are significantly worse than the best. $\pm$ is half the width of the 95\% confidence interval.}
\label{tab:iwslt_sig}
\end{table*}
\begin{table*}[ht!]
\centering
\footnotesize
\setlength{\tabcolsep}{3pt}
\begin{tabular}{lccc|ccccc}
\toprule
& \multicolumn{3}{c|}{\textbf{MuST-SHE} PA (\%)} & \multicolumn{5}{c}{\textbf{ContraProST} PA (\%)} \\
\cmidrule(lr){2-4}\cmidrule(lr){5-9}
\textbf{Model} & \textbf{es} & \textbf{fr} & \textbf{it} & \textbf{Emo.\ Prosody} & \textbf{Intonation} & \textbf{Prag.\ Prosody} & \textbf{Pros.\ Breaks} & \textbf{Sent.\ Stress} \\
\midrule
COMET-PARTIAL & 52.4 & 55.7$\uparrow$ & 51.8 & 50.0 & 50.0 & 50.0 & 50.0 & 50.0 \\
COMETKiwi & 61.5$\uparrow$ & 60.6$\uparrow$ & 55.2$\uparrow$ & 50.0 & 50.0 & 50.0 & 50.0 & 50.0 \\
SpeechQE & 37.0$\downarrow$ & 36.0$\downarrow$ & 31.0$\downarrow$ & 27.3$\downarrow$ & 32.0$\downarrow$ & 31.0$\downarrow$ & 23.9$\downarrow$ & 27.7$\downarrow$ \\
BLASER & 52.0 & 51.6 & 51.7 & 49.8 & 56.0$\uparrow$ & 49.8 & 50.5 & 50.3 \\
\midrule
COMETKiwi$_{\text{RoBERTa}}^{\text{WMT}}$ & 51.2 & 52.8 & 50.1 & 50.0 & 50.0 & 50.0 & 50.0 & 50.0 \\
COMETKiwi$_{\text{RoBERTa}}^{\text{IWSLT}}$ & 52.8 & 49.1 & 52.0 & 50.0 & 50.0 & 50.0 & 50.0 & 50.0 \\
SpeechCOMET\textsubscript{SONAR} & 53.1 & 50.9 & 50.9 & 50.1 & 49.9 & 50.0 & 50.1 & 49.8 \\
SpeechCOMET\textsubscript{Whisper} & 51.4 & 50.7 & 52.8 & 50.1 & 50.2 & 50.0 & 49.9 & 49.9 \\
SpeechCOMET & 55.6$\uparrow$ & 52.6 & 52.4 & 50.0 & 49.8 & 49.9 & 50.0 & 49.9 \\
SpeechCOMET$^{\text{XL}}$ & 54.6$\uparrow$ & 58.0$\uparrow$ & 51.1 & 50.0 & 50.1 & 50.0 & 50.0 & 50.0 \\
\midrule
SpeechLLM (\textsc{Text}) & 23.7 & 17.3$\downarrow$ & 16.2$\downarrow$ & 25.3 & 27.3 & 26.8 & 20.0$\downarrow$ & 22.9$\downarrow$ \\
\quad +FT & 14.0$\downarrow$ & 13.8$\downarrow$ & 15.4$\downarrow$ & 18.4$\downarrow$ & 22.0$\downarrow$ & 18.1$\downarrow$ & 13.9$\downarrow$ & 18.4$\downarrow$ \\
SpeechLLM (\textsc{Speech}) & 38.0 & 32.3 & 32.4 & 32.3$\downarrow$ & 41.2$\downarrow$ & 31.9$\downarrow$ & 23.9$\downarrow$ & 27.5$\downarrow$ \\
\quad +FT & 35.9$\downarrow$ & 15.0 & 20.7 & 20.6$\downarrow$ & 24.9$\downarrow$ & 17.4$\downarrow$ & 15.5$\downarrow$ & 19.2$\downarrow$ \\
SpeechLLM (\textsc{Sp.+Txt}) & 43.1 & 37.1 & 37.8 & 30.6$\downarrow$ & 36.4$\downarrow$ & 32.1$\downarrow$ & 23.4$\downarrow$ & 27.7$\downarrow$ \\
\quad +FT & 8.2 & 6.3 & 6.8 & 17.1$\downarrow$ & 19.7$\downarrow$ & 12.3$\downarrow$ & 14.2$\downarrow$ & 15.6$\downarrow$ \\
\bottomrule
\end{tabular}
\caption{MuST-SHE and ContraProST results across all models. $\uparrow$/$\downarrow$ indicate significantly ($p<0.05$) above/below chance under a permutation test whose null resamples synthetic correct/wrong scores from the model's own pooled values for that cell.}
\label{tab:mustshe_contraprost_full}
\end{table*}
\begin{table}[ht!]
\centering
\footnotesize
\setlength{\tabcolsep}{4pt}
\begin{tabular}{ll|rc}
\toprule
 \multicolumn{2}{r}{\textbf{Segment $\tau_b$ (\%)}} & \textbf{real src} & \textbf{$\Delta$ random} \\
\midrule
\multicolumn{4}{l}{\textit{Baselines}} \\
\midrule
\multirow{2}{*}{\textsc{Text}} & \textsc{COMET-Part.} & \cellcolor[RGB]{255,255,255}11.9 & \cellcolor[RGB]{249,251,253}-4.3$\pm$3.9 \\
 & \textsc{COMETKiwi} & \cellcolor[RGB]{212,236,236}34.6 & \cellcolor[RGB]{194,215,235}-24.9$\pm$4.5$^*$ \\
\multirow{2}{*}{\textsc{Speech}} & \textsc{SpeechQE} & \cellcolor[RGB]{221,240,240}29.8 & \cellcolor[RGB]{201,219,237}-22.5$\pm$5.4$^*$ \\
 & \textsc{BLASER} & \cellcolor[RGB]{231,244,244}24.4 & \cellcolor[RGB]{220,232,243}-15.4$\pm$4.8$^*$ \\
\midrule
\multicolumn{4}{l}{\textit{SpeechCOMET}} \\
\midrule
\multirow{2}{*}{\textsc{Text}} & COMETKiwi$_{\text{RoBERTa}}^{\text{WMT}}$ & \cellcolor[RGB]{234,246,245}23.1 & \cellcolor[RGB]{231,239,247}-11.1$\pm$4.5$^*$ \\
 & COMETKiwi$_{\text{RoBERTa}}^{\text{IWSLT}}$ & \cellcolor[RGB]{234,246,246}22.8 & \cellcolor[RGB]{229,238,246}-12.0$\pm$3.8$^*$ \\
\multirow{2}{*}{\textsc{Speech}} & SpeechCOMET\textsubscript{SONAR} & \cellcolor[RGB]{239,248,248}20.1 & \cellcolor[RGB]{253,253,254}-3.1$\pm$3.0 \\
 & SpeechCOMET\textsubscript{Whisper} & \cellcolor[RGB]{244,250,250}17.8 & \cellcolor[RGB]{255,255,255}-2.4$\pm$3.1 \\
\textsc{Sp.+Txt} & SpeechCOMET & \cellcolor[RGB]{228,243,243}25.8 & \cellcolor[RGB]{213,228,241}-17.8$\pm$4.5$^*$ \\
\textsc{Sp.+Txt} & SpeechCOMET$^{\text{XL}}$ & \cellcolor[RGB]{212,236,236}34.4 & \cellcolor[RGB]{180,206,231}-30.2$\pm$4.7$^*$ \\
\midrule
\multicolumn{4}{l}{\textit{SpeechLLM}} \\
\midrule
\multirow{2}{*}{\textsc{Text}} & SpeechLLM & \cellcolor[RGB]{201,232,231}40.4 & \cellcolor[RGB]{183,208,232}-29.2$\pm$18.6 \\
 & \quad+FT & \cellcolor[RGB]{188,226,226}47.1 & \cellcolor[RGB]{157,191,223}-38.6$\pm$34.6$^*$ \\
\multirow{2}{*}{\textsc{Speech}} & SpeechLLM & \cellcolor[RGB]{217,238,238}32.0 & \cellcolor[RGB]{186,209,232}-28.1$\pm$10.0$^*$ \\
 & \quad+FT & \cellcolor[RGB]{202,232,231}40.1 & \cellcolor[RGB]{193,214,235}-25.5$\pm$14.0$^*$ \\
\multirow{2}{*}{\textsc{Sp.+Txt}} & SpeechLLM & \cellcolor[RGB]{205,233,233}38.1 & \cellcolor[RGB]{161,193,224}-37.4$\pm$17.6$^*$ \\
 & \quad+FT & \cellcolor[RGB]{183,224,223}49.9 & \cellcolor[RGB]{156,190,223}-39.2$\pm$26.1$^*$ \\
\bottomrule
\end{tabular}
\caption{Effect of replacing the source input with a randomly mismatched one. Scores are averaged over de and zh. $\Delta$ = random$-$real: large negative values indicate the model uses the source signal. $\pm$ is half the width of the 95\% confidence interval; $^*$ indicates a significant difference ($p<0.05$) between real-source and shuffled-source scores; no asterisk means the two are statistically indistinguishable.}\label{tab:random_src_sig}
\vspace{-2mm}
\end{table}

\subsection{SpeechLLM Prompt Sensitivity}

\cref{tab:speechllm_prompt} reports MuST-SHE and ContraProST PA for SpeechLLM with and without a speech-aware prompt across all input modalities. All values fall below 50\%, reflecting the tie-heavy prediction pattern described in \cref{sec:results}.

\subsection{Significance Analysis}\label{subsec:significance}

To assess significance, we use paired bootstrap resampling ($p\geq0.05$ tie with the best model) for the segment-level correlation scores, and a permutation test against each model's own pooled predictions ($p<0.05$ for above/below chance) for the contrastive pairwise-accuracy scores.

For the segment-level correlation comparisons, SpeechLLM (+FT, speech+text) achieves the highest $\tau_b$, and several other models, including the off-the-shelf COMETKiwi baseline, are not significantly different from it (\cref{tab:iwslt_sig}). For the contrastive pairwise-accuracy scores,  \cref{tab:mustshe_contraprost_full} shows that the COMET- and SpeechCOMET-family metrics are statistically indistinguishable from chance on ContraProST. On MuST-SHE, only COMETKiwi consistently scores statistically above chance, and only at 55–62\% PA, too low for reliable gender-agreement evaluation. SpeechQE and SpeechLLM score significantly below chance on both benchmarks, driven by the tie-heavy prediction pattern described in Section 4.

We also run a paired bootstrap significance test on the real-vs-shuffled-source comparison (\cref{tab:random_src_sig}). Most metrics show a significant drop when the source is shuffled, confirming reliance on the source signal, with the notable exception of SpeechCOMET-SONAR and SpeechCOMET-Whisper, whose real and shuffled scores are statistically indistinguishable, confirming our claim that these speech-only variants largely ignore the source. All speech and speech+text SpeechLLM variants show significant drops, supporting our conclusion that SpeechLLMs condition on the source.

\section{Analysis Details}\label{app:analysis_details}
\subsection{Probing Setup}\label{app:probing_setup}

  \begin{table}[ht!]
  \centering
  \small
  \setlength{\tabcolsep}{2pt} 
  \begin{tabular}{@{}ll@{}}
  \toprule
  \textbf{Hyperparameter} & \textbf{Value} \\
  \midrule
  \multicolumn{2}{@{}l}{\textit{Architecture}} \\[2pt]
  Classifier            & MLP \\
  Input dimension       & 768 for (speech)COMET         \\
                        & 3584 for Qwen2.5-Omni \\
  Hidden dimension      & 256 \\
  Activation            & ReLU \\
  Output layer          & Linear ($\to$ \#classes) \\
  Dropout               & 0.1 \\
  \midrule
  \multicolumn{2}{@{}l}{\textit{Optimisation}} \\[2pt]
  Optimizer             & Adam \\
  Learning rate         & $10^{-3}$ by default \\
                        & $3\times10^{-4}$ (Qwen2.5-Omni) \\
  Batch size            & 256 \\
  \midrule
  \multicolumn{2}{@{}l}{\textit{Data}} \\[2pt]
  Linguistic feat.     & IWSLT 2026 Metric (English source) \\
                             & training data sampled with 10\% \\
  Speaker gender         & MuST-SHE v1.2 (category-1 only) \\
                             & 50/50 stratified split for train/test \\
  Intonation             & ContraProST ``prosodic breaks'' subset \\
                             & 80/20 stratified split for train/test\\
  Emotion                & ContraProST ``emotional prosody'' subset \\
                             & 80/20 stratified split for train/test   \\
  \bottomrule
  \end{tabular}
  \caption{Probing classifier configuration.}
  \label{tab:probe-config}
  \end{table}

Training details of the probing classifiers are presented in \cref{tab:probe-config}.

\begin{figure*}[t!]
  \centering
\includegraphics[width=\textwidth]{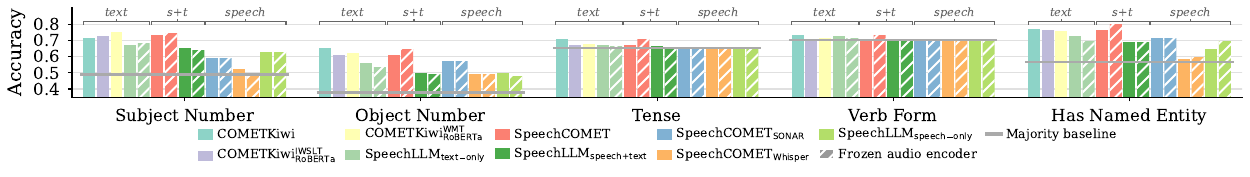}
  \caption{Probing accuracy for linguistic features on the IWSLT 2026 Metrics dataset, grouped by input modality (text-only, speech+text, speech-only).
}
\label{fig:linguistics_feat_probing_extended}
\end{figure*}
\begin{figure*}[ht!]
  \centering
\includegraphics[width=\textwidth]{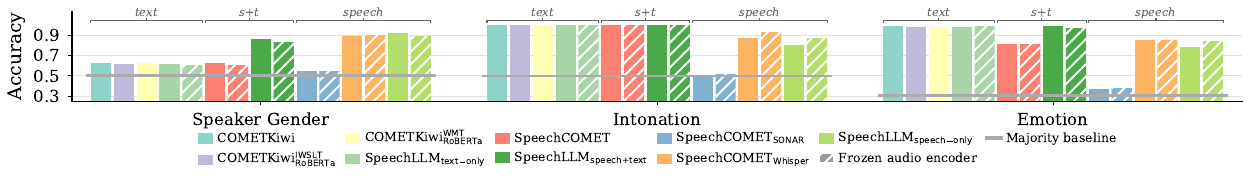}
  \caption{Probing accuracy for acoustic features on MuST-SHE (speaker gender) and ContraProST (intonation, emotion).
  Text models are unable to recover \textit{speaker gender} accurately, so the speech-only systems must rely on acoustic cues. 
  For \textit{intonation} and \textit{emotion}, the dataset's source transcripts expose the feature (trailing punctuation and typographic prosody markers respectively), which explains the high text-only accuracy on these two.
}
\label{fig:acoustic_feat_probing_extended}
\end{figure*}

\subsubsection{Linguistic Feature Probing Data}
The linguistic probing dataset is derived from the IWSLT 2026 Metrics Shared Task dataset, which pairs English and Czech source speech and text with MT hypotheses and human quality scores across several language pairs. As the morphological tagger is English-only, we restrict probing to English sources, yielding 31,965 training and 5,556 development examples.

We annotate the source side with spaCy's \texttt{en\_core\_web\_trf} model and extract five sentence-level features. ``Verb Form'' and ``Tense'' are read from the \texttt{ROOT} token of the dependency parse. ``Subject Number'' and ``Object Number'' give the grammatical number (singular/plural) of the first \texttt{nsubj} and \texttt{obj} dependents. ``Has Named Entity'' indicates whether spaCy recognises any entity span, covering both proper entities (e.g., PERSON, ORG, GPE) and numeric or temporal expressions (e.g.\ DATE, CARDINAL, MONEY). Sentences for which a label is undefined (e.g., no object for Object Number) are excluded from that feature only.

\subsubsection{Acoustic Feature Probing Data}
\paragraph{Speaker Gender.}
Derived from MuST-SHE v1.2, which provides the English source, a reference translation, a gender label (He/She), and a category label for three directions (En→Es/Fr/It). We keep only category-1 segments, where gender is lexically recoverable from the utterance, and deduplicate by (talk ID, source text, speaker). Stratified splitting yields two 526-segment halves (He: 265, She: 261).

\paragraph{Intonation.}
Constructed from the ContraProST Intonation subset: 173 English sentences each recorded as a declarative (``.'') and a yes/no question (``?''), for 346 files. We deduplicate by audio path across the three target-language files (En→De/Es/Ja). Splitting at the sentence level (both recordings in the same partition), an 80/20 split gives 276 training (138/138) and 70 development (35/35) examples. Note that the statement/question distinction surfaces in the source text as \textit{trailing punctuation}, which is why text-based models reach near-perfect accuracy.

\paragraph{Emotion.}
Also from ContraProST, using the Emotional Prosody subset, where each sentence is recorded under multiple emotions. Of the six original classes we keep the four most frequent (Neutral, Angry, Happy, Surprised), as the other two classes contain too few samples. Splitting at the sentence level, an 80/20 split gives 964 training (Neutral: 301, Angry: 292, Happy: 251, Surprised: 120) and 247 development (76/75/64/32) examples. As with intonation, the transcripts carry class-discriminative typography: neutral utterances are plain text, while angry, happy, and surprised ones use asterisks (\texttt{*word*}), underscores (\texttt{\_word\_}), capitalisation, and repeated punctuation (e.g.\ ?!?!), again allowing high text-based accuracy.

\subsection{Additional Probing Results}\label{app:additional_probing_results}

The probing results for the full set of models are presented in \cref{fig:linguistics_feat_probing_extended} and \cref{fig:acoustic_feat_probing_extended}, corresponding to linguistic and acoustic features respectively.

\subsection{Source Modality Ablation for SpeechCOMET (Speech+Text)}\label{app:source_ablation}

\begin{table}[ht!]
\centering
\footnotesize
\setlength{\tabcolsep}{4pt}
\begin{tabular}{l|rrrr}
\toprule
\textbf{Model} & \textbf{real src} & \textbf{$\Delta$ both} & \textbf{$\Delta$ audio} & \textbf{$\Delta$ text} \\
\cmidrule(lr){2-2}\cmidrule(lr){3-5}
\multicolumn{5}{l}{\textit{Segment $\tau_b$ avg (\%) over de \& zh}} \\
\midrule
SpeechCOMET & \cellcolor[RGB]{255,255,255}25.8 & \cellcolor[RGB]{220,230,245}-17.8 & \cellcolor[RGB]{251,252,253}-1.4 & \cellcolor[RGB]{208,224,240}-17.2 \\
SpeechCOMET$^{\text{XL}}$ & \cellcolor[RGB]{183,224,223}34.4 & \cellcolor[RGB]{173,201,228}-30.2 & \cellcolor[RGB]{254,254,254}-0.3 & \cellcolor[RGB]{172,200,228}-30.6 \\
\bottomrule
\end{tabular}
\caption{Source-modality ablation for speech+text SpeechCOMET models. $\Delta$ audio only: audio replaced with mismatched audio, text kept real. $\Delta$ text only: text replaced with mismatched text, audio kept real. $\Delta$ both: both modalities replaced. Scores are segment $\tau_b$ averaged over de and zh.}\label{tab:source_ablation}
\end{table}

\cref{tab:source_ablation} disentangles the contribution of each input modality for the two speech+text SpeechCOMET models by separately replacing audio or text with a randomly mismatched source. For both models, replacing only the text source reproduces nearly the full drop observed when both modalities are shuffled ($-17.2$ pp for SpeechCOMET, $-30.6$ pp for SpeechCOMET$^{\text{XL}}$), while replacing only the audio has a negligible effect ($-1.4$ and $-0.3$ pp respectively). The pattern is consistent across both encoder sizes, confirming that the multimodal models condition almost exclusively on the text source.

\end{document}